\documentclass[journal]{IEEEtran}

\usepackage{amsmath,amssymb,amsfonts}
\usepackage{graphicx}
\usepackage{dblfloatfix}
\usepackage{placeins}
\usepackage{caption}
\usepackage{booktabs}
\usepackage{array}
\usepackage{multirow}
\usepackage{cite}
\usepackage{url}
\usepackage{siunitx}
\usepackage{algorithm}
\usepackage{algpseudocode}

\title{UnMixMamba: Unmixing Guided State Space Subpixel Mapping of Crops from MODIS Time Series}

\author{Motasem Alkayid, Yimin Zhu, Naser El-Sheimy,~\IEEEmembership{Member,~IEEE}, Lincoln Linlin Xu,~\IEEEmembership{Member,~IEEE}
\thanks{This work was supported by the Natural Sciences and Engineering Research Council of Canada (NSERC) under Grant RGPIN-2019-06744.}
\thanks{Yimin Zhu, Naser El-Sheimy, and Lincoln Linlin Xu are with the Department of Geomatics Engineering, University of Calgary, Canada (email: \{yimin.zhu, elsheimy, lincoln.xu\}@ucalgary.ca). Corresponding author: Lincoln Linlin Xu.}
\thanks{Motasem Alkayid is with the Department of Geomatics Engineering, University of Calgary, Canada, and also with the Department of Geography, Faculty of Arts, The University of Jordan, Amman, Jordan (email: motasem.alkayid@ucalgary.ca).}
}

\begin{document}
\maketitle

\begin{abstract}
Subpixel mapping (SPM) of crop types from \SI{250}{m} Moderate Resolution Imaging Spectroradiometer (MODIS) time series to field-scale Agriculture and Agri-Food Canada (AAFC) labels supports acreage estimation and agricultural monitoring. Three difficulties remain: (i)~spectral--temporal coupling and mixed pixels that break direct coarse-to-fine classifiers; (ii)~intra-class endmember variability that a single spectrum per crop cannot capture; and (iii)~uniform refinement on the high-resolution (HR) grid, which treats every subpixel the same, so most of the effort goes to easy pixels while errors remain in the minority class. UnMixMamba addresses them as follows. For~(i), a temporal--spectral--spatial decoupled (TSSD) factorization estimates simplex fractions and drives AAFC classification from the same mixture field, instead of mapping low-resolution (LR) observations directly to HR logits. For~(ii), \(\times 8\) abundance super-resolution and a learnable per-crop variant endmember bank reconstruct the MODIS cube at LR and HR under a linear mixing model. For~(iii), the predicted crop fractions decide where a fixed Mamba token budget is spent, so HR refinement focuses on the minority class instead of the full grid. On Alberta 2015 under a fixed spatial-block protocol, UnMixMamba reaches OA 0.8397, mIoU 0.6640, AA 0.7837, and Kappa 0.7715, ahead of ST-DRes, ESPCN, FineCrop, AgriFM, and classical SPM baselines. The task is closed-set four-crop mapping inside a known agricultural mask.
\end{abstract}


\section{Introduction}
\label{sec:intro}

Field-scale crop maps are used to estimate planted area, yield, rotation, and food-security statistics. In Canada, Agriculture and Agri-Food Canada (AAFC) produces an annual crop inventory at about \SI{30}{m}, the usual reference product for the Prairie provinces \cite{aafc_open_data,aafc_gee,valizadeh2020canada}.
However, operational production of such maps depends on cloud-free, high-resolution imagery throughout the growing season, which is not guaranteed in every region and year. The Moderate Resolution Imaging Spectroradiometer (MODIS) archive spans two decades with near-daily coverage; its \SI{16}{day} MOD13Q1 composites capture the full phenological trajectory at \SI{250}{m}. The primary constraint lies in spatial resolution: an individual \SI{250}{m} MODIS pixel typically covers one or more Prairie fields, so the coarse signal is a mixture of multiple crop types rather than a pure spectral signature.

Subpixel mapping (SPM), addresses this challenge by predicting fine-resolution class labels from coarse observations while maintaining consistency with the estimated subpixel class proportions \cite{atkinson1997mapping}. Classical SPM estimates per-class abundances first and then allocates hard labels using spatial attraction or temporal continuity priors \cite{shi2014sthnn}. These allocators are interpretable and mixture-aware but rely on hand-designed priors and degrade when fractions are noisy. Recent deep models map the coarse low-resolution (LR) cube directly to high-resolution (HR) classes \cite{stdres2022}. Accuracy rises, but reconstructed spectra and abundances are rarely required to agree across scales. An end-to-end model is still needed that keeps the mixture constraint of classical SPM, learns the allocation stage, and forces abundances and reconstructed spectra to stay consistent between the coarse and fine grids.

In the MODIS to AAFC setting, three difficulties dominate:
First~\emph{(i)}, spectral--temporal coupling and mixed pixels: The MODIS data cube comprises $23$ 16-day composites with six spectral bands per date. Flattening this structure discards phenological trajectories, and mixed \SI{250}{m} pixels dominate field boundaries. Standard deep learning pipelines often predict HR class scores directly and omit explicit mixture constraints \cite{ling2019super}.
Second~\emph{(ii)}, intra-class endmember variability: A single crop type can span a wide range of spectra because of planting date, soil, and moisture, so a fixed single endmember is a poor model of Prairie fields \cite{somers2011endmember}.
Third~\emph{(iii)}, uniform HR refinement: A single \(512\times512\) HR patch has more than \(2.6\times10^{5}\) subpixels. Most refinement methods treat all of them the same and ignore the crop fractions already estimated for each pixel. Since most subpixels lie inside large, uniform fields of common crops, this spends most of the effort where labels are already easy, while the minority class, where most errors remain, gets little of it.

We develop UnMixMamba, an end-to-end mixture-constrained network for MODIS to AAFC crop SPM. Challenges~(i)--(iii) are handled by a shared abundance field, variant-endmember reconstruction, and abundance-guided HR refinement, respectively.

\subsection{Contributions}
\label{sec:contributions}
The main contributions of this article, aligned with the aforementioned challenges~(i)--(iii), are summarized as follows. They are implemented through building blocks C1--C3; Table~\ref{tab:novelty_audit} positions each block against the closest prior work.
\begin{enumerate}
    \item \emph{Mixture-Constrained TSSD Abundance SPM:} To address the issue of spectral--temporal coupling and mixed pixels, we propose a temporal--spectral--spatial decoupled (TSSD) factorization (C1). This module estimates simplex abundances from the coarse MODIS data cube and drives the high-resolution classification from a unified mixture field. Consequently, the subpixel mapping process is explicitly constrained by a fractional representation, avoiding the pitfalls of direct low-resolution to high-resolution logit mapping.
    
    \item \emph{Variant-Endmember Dual-Scale Reconstruction:} To capture intra-class endmember variability, we introduce a learnable per-crop variant bank coupled with an abundance super-resolution module (C2). The MODIS cubes are reconstructed at both low and high resolutions under a linear mixing model. This design lets spectral variability be absorbed within each crop class rather than being attributed across different classes.
    
    \item \emph{Abundance-Guided HR Refinement:} Instead of refining every subpixel, we use the predicted crop fractions (abundances) to decide where to refine (C3). A fixed number of HR tokens is shared among the crops based on their predicted fractions, with a guaranteed minimum share for the hardest classes, and only these tokens are refined by crop-specific Mamba blocks. In this way, the same abundance field that drives reconstruction and classification also decides where the computation goes: refinement runs on about 11\% of the HR grid and focuses on the classes with the most errors.
\end{enumerate}

The remainder of this article is organized as follows. Section~\ref{sec:related} reviews related work. Section~\ref{sec:dataset} describes the study area and data, and Section~\ref{sec:method} presents UnMixMamba. Section~\ref{sec:exp} details the experimental setup, Section~\ref{sec:results} reports and discusses the results, and Section~\ref{sec:conclusion} concludes the article.

\section{Related Work}
\label{sec:related}

We organize prior work by the three challenges in Section~\ref{sec:contributions}: SPM and mixed pixels (Section~\ref{sec:rw_spm}), unmixing and endmember variability (Section~\ref{sec:rw_unmix}), and abundance-guided HR refinement with efficient sequence models (Section~\ref{sec:rw_mamba}), followed by the AAFC label context (Section~\ref{sec:rw_aafc}). Table~\ref{tab:novelty_audit} maps blocks C1--C3 to the nearest prior methods.

\begin{table*}[!tp]
\caption{Building blocks of UnMixMamba (C1--C3, implementing Contributions~1--3) versus the closest prior work.}
\label{tab:novelty_audit}
\centering
\footnotesize
\setlength{\tabcolsep}{4pt}
\renewcommand{\arraystretch}{1.35}
\setlength{\extrarowheight}{4pt}
\begin{tabular}{@{}p{0.11\textwidth}p{0.26\textwidth}p{0.295\textwidth}p{0.295\textwidth}@{}}
\toprule
\textbf{Block} & \textbf{Closest prior art} & \textbf{What they already do} & \textbf{Role inside UnMixMamba} \\
\midrule
Task & SPMCNN-F \cite{he2021spmcnn}; ST-DRes \cite{stdres2022}; PHAT \cite{phat2025}; STSPM\_MO \cite{stspm2024} & Direct LR$\rightarrow$HR crop logits; phenology attention; multiobjective allocation from \emph{given} abundances & Same task (MODIS crop SPM), but abundances are learned end-to-end with reconstruction \\[4pt]
C1 & CropSTS \cite{cropsts2025}; EnMAP dual stream \cite{enmap2025}; FineCrop \cite{finecrop2025}; TAS2B-Net \cite{tas2b2025} & Decouple streams to predict \emph{class logits} & Emit \emph{simplex abundances} that start the shared fraction loop (Contribution~1) \\[4pt]
C2 & MESMA \cite{mesma1998}; SSAF-Net \cite{ssafnet2025}; EOT-Net \cite{eotnet2025}; GLSV-Net \cite{glsv2025}; Hu \emph{et al.} \cite{snlsr2024}; SMILE \cite{smile2025}; UnmixingSR \cite{yu2024unmixingsr}; Jia \emph{et al.} \cite{jia2025tgrs_cropland_sr} & HSI variability/unmixing; or Sentinel-2 binary cropland SR from simulated fractions & Variant bundles + LR/HR recon + abundance SR under real multi-crop AAFC labels (Contribution~2) \\[4pt]
C3 & SITSMamba \cite{qin2024sitsmamba}; STSMamba \cite{stsmamba2025}; KAMamba \cite{kamamba2025}; S2Mamba \cite{s2mamba2025} & Temporal classification/change detection; sparse MODIS tokens & Use predicted crop fractions to decide where to refine on the HR grid (Contribution~3) \\
\bottomrule
\end{tabular}
\end{table*}

\subsection{Classical and Deep Subpixel Mapping}
\label{sec:rw_spm}

Subpixel mapping (super-resolution mapping) converts coarse fractions into a hard fine-resolution label map \cite{atkinson1997mapping,atkinson2005psa}. The classical pipeline is two-stage: estimate per-class abundances, then allocate subpixels under a spatial dependence or attraction prior \cite{mertens2006saps,shi2014sthnn}. Pixel swapping, Hopfield networks, radial-basis interpolation, and related metaheuristics remain useful references because they enforce the fraction constraint without training \cite{atkinson2005psa,wang2014rbf,ling2010sa}. Their limitation is structural: the allocation prior is fixed and local, so accuracy is capped by the quality of the input fractions and by how well generic smoothness matches Prairie field geometry. When fractions are noisy, as they are for spectrally similar cereals at \SI{250}{m}, these allocators fragment minority classes into speckle.

Deep SPM replaces the hand-designed allocator with a learned LR$\rightarrow$HR map. SFRNet-MLFF uses multilevel spatial feedback on hyperspectral imagery \cite{sfrnet2024}; SSNET jointly models unmixing and SPM to reduce error propagation from a separate unmixing stage \cite{ssnet2021}; STSPM\_MO allocates from abundance, spatial, and temporal objectives, optionally with historical maps \cite{stspm2024}. Learning-based super-resolution land-cover mapping similarly trains a CNN to recover fine labels from coarse fractions \cite{ling2019super}. These methods are developed mainly on hyperspectral or simulated fraction inputs. They do not learn a MODIS temporal--spectral--spatial unmixing front end, a variant endmember bank, or an HR Mamba readout under AAFC crop labels.

For MODIS crop mapping, SPMCNN-F maps coarse MODIS stacks directly to HR crop labels \cite{he2021spmcnn}. ST-DRes is the strongest published end-to-end MODIS crop SPM system: spatial--temporal depth-wise residual blocks with PixelShuffle upsampling, including spatial and temporal transfer experiments \cite{stdres2022}. In that study, UNet is adapted for SPM by bilinear upsampling before the classifier. PHAT uses phenology-aware attention for large-area mixed-pixel estimation \cite{phat2025}, but it does not perform field-scale AAFC SPM with a mixture-reconstruction loop. Note also that optical image super-resolution reconstructs HR radiance or vegetation indices, whereas SPM allocates hard class labels under a fraction constraint. UnMixMamba remains in MODIS crop SPM with end-to-end unmixing and allocation under AAFC labels: TSSD abundances, LR/HR spectral reconstruction, and HR classification share one mixture field.

\subsection{Spectral Unmixing, Endmember Variability, and Decoupled Encoders}
\label{sec:rw_unmix}

Spectral unmixing recovers the fractional contribution of each class inside a mixed pixel and is the usual formulation of coarse-resolution crop analysis \cite{hu_review2025}. For MODIS, temporal mixture analysis exploits the fact that crops follow distinct vegetation-index trajectories, so a mixture that is ambiguous on any single date can still be resolved over the season \cite{tao2023modis}. ST-Bdip refines LR abundances with a U-Net under a temporal mixing model but stops at the coarse scale and does not allocate HR labels \cite{stbdip2023}.

A single reference spectrum per crop cannot represent planting-date, cultivar, soil, and moisture spread. Multiple-endmember spectral mixture analysis (MESMA) \cite{mesma1998}, endmember bundles \cite{bundle2025}, and learnable variability models \cite{borsoi2021variability} address this in the unmixing literature; recent TGRS networks include SSAF-Net \cite{ssafnet2025}, EOT-Net \cite{eotnet2025}, MAHUM \cite{mahim2023}, and global-to-local scaled-variability models \cite{glsv2025}. These systems are evaluated mainly on hyperspectral unmixing metrics, not on multi-crop AAFC SPM. Abundance-domain super-resolution likewise argues for upsampling fractions rather than radiance \cite{yu2024unmixingsr,snlsr2024,ucnn2021,smile2025}, but remains largely an HSI benchmark; Jia \emph{et al.} \cite{jia2025tgrs_cropland_sr} treat Sentinel-2 binary cropland with simulated fractions.

Decoupled spatiotemporal encoders for crop time series \cite{cropsts2025,enmap2025,finecrop2025,tas2b2025} improve phenological discrimination, yet they emit class logits. Here, C1 applies the same decoupling to \emph{simplex abundance} estimation; C2 adds \(\times 8\) abundance super-resolution and LR/HR reconstruction under AAFC labels with a learnable variant bank.

\subsection{Satellite Image Time Series, State-Space Models, and Sparse Mamba}
\label{sec:rw_mamba}

At native spatial resolution, crop mapping from satellite image time series (SITS) is well studied. TempCNN applies 1-D temporal convolutions to phenological profiles \cite{tempcnn2019}; U-TAE combines a U-Net spatial backbone with a temporal attention encoder and introduced the PASTIS agricultural benchmark \cite{utae2021}; TSViT factorizes spatial and temporal attention for SITS segmentation \cite{tsvit2023}. These models assume that each pixel is already at field scale (typically Sentinel-2). They do not solve coarse-to-fine MODIS SPM, and a quadratic attention encoder over a \(512\times512\) HR grid is not a practical refinement module for that task.

State-space models, and Mamba in particular, provide linear-time sequence modeling as an alternative to attention \cite{mamba2024,mamba_survey_2026,mamba_arch_2026}. Remote sensing adaptations include SITSMamba for crop classification from SITS \cite{qin2024sitsmamba}, GWSC-SegMamba for multi-temporal agricultural segmentation \cite{gwscsegmamba2025}, and spatial--spectral Mamba classifiers \cite{s2mamba2025,mambassfn2025}. On MODIS specifically, STSMamba uses a temporal grouped stem and sparse deformable Mamba sequences for land-cover classification \cite{stsmamba2025}; KAMamba extends the same sparse backbone to joint change detection and classification \cite{kamamba2025}. Dynamic or saliency-driven scans likewise sparsify vision Mamba \cite{damamba2025,vfmamba2025}. None of these methods allocate HR crop labels from mixture abundances.

UnMixMamba adapts Mamba using a fixed token budget derived from predicted crop abundances, prioritizing refinement capacity for minority classes and high-abundance regions rather than expending it on homogeneous interiors across a $512 \times 512$ map ($> 2.6 \times 10^5$ tokens). The core contribution lies in this abundance-guided selection rule rather than the underlying choice of sequence operator (Section~\ref{sec:c3}).

\subsection{AAFC Crop Inventory and Confusable Crops}
\label{sec:rw_aafc}

The AAFC annual crop inventory is the operational \SI{30}{m} national product, produced with a decision-tree classifier on multi-temporal optical and radar imagery and validated over the Prairies \cite{fisette2013aafc,aafc_open_data,aafc_gee,valizadeh2020canada}. It is a strong HR supervision target, but it also imposes AAFC class definitions and class imbalance on any model trained against it. Cloud-free HR coverage during the growing season is not guaranteed every year, which is the operational reason to recover field-scale labels from the MODIS archive. Foundation crop models such as AgriFM \cite{agrifm2025} and fine-grained remappers such as FineCrop \cite{finecrop2025} learn transferable representations at native sensor resolution; they do not perform coarse-to-fine MODIS SPM.

Conventional approaches to class imbalance (such as loss reweighting and label smoothing) \cite{imbalance2025} fail to exploit fine-grained off-diagonal error geometry. Although the spectral confusion between wheat and barley persists at higher resolutions \cite{wheatbarley2025}, the AAFC residuals demonstrate that the actual confusion is tripartite: true Barley is simultaneously misallocated to Spring Wheat and Canola (Section~\ref{sec:results}). We formulate this dynamic as the \emph{three-way Barley hub}, wherein Barley forms the intersection of a cereal-cereal axis (Spring Wheat and Barley) and an oilseed-cereal axis (Barley and Canola). Consequently, UnMixMamba targets both axes: two pairwise logit-margin losses enforce class boundaries between Spring Wheat and Barley as well as Barley and Canola, while C3 introduces a \emph{hard-class floor} that reserves at least 3\% of the high-resolution grid for underrepresented, high-confusion classes (Section~\ref{sec:c3}).

\section{Study Area and Dataset}
\label{sec:dataset}
\subsection{Study Region and Labels}
The study area is the agricultural extent of Alberta, Canada, for crop year 2015. Alberta is part of the Canadian Prairie ecozone, the country's dominant cereal- and oilseed-producing region, and its fields are large and geometrically regular, which makes it a representative and demanding testbed for \SI{250}{m} SPM: field widths are frequently smaller than a MODIS pixel, so mixing occurs along essentially every field boundary. High-resolution supervision uses the AAFC annual crop inventory (approximately \SI{30}{m}), which we resample onto an aligned \SI{31.25}{m} grid so that the HR and LR grids share an integer scale factor (\(8\times\SI{31.25}{m}=\SI{250}{m}\)). This integer alignment provides numerical grid correspondence for block pooling and cross-scale reconstruction, but it does \emph{not} eliminate residual MODIS geolocation error, resampling uncertainty, or the mismatch between average pooling and the true MODIS point spread function (PSF). We therefore treat \(\times 8\) alignment as a convenient geometric convention rather than a claim of perfect registration.

\subsection{Input Data}
\begin{figure}[!htbp]
\centering
\vspace{-4pt}
\includegraphics[width=\columnwidth]{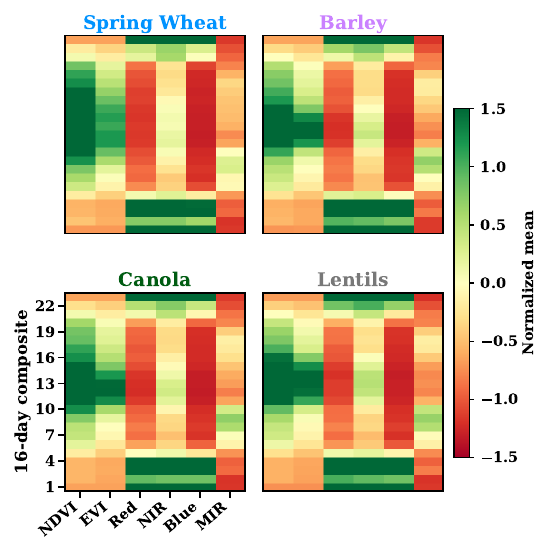}
\vspace{-4pt}
\caption{Per class mean MODIS input signatures on Alberta 2015 training pixels: heatmaps over 23 sixteen-day composites (rows) and six bands per date (NDVI, EVI, Red, NIR, Blue, MIR). Values are z-scored within each crop panel for side-by-side comparison. At \SI{250}{m}, phenological curves overlap and spectral--temporal differences are small. Spring Wheat and Barley are hard to separate (near-identical NDVI trajectories, only subtle Red/NIR/MIR separation); Canola shows a clearer mid-season green-up; Lentils has a shorter growing-season envelope.}
\label{fig:modis_signatures}
\end{figure}

Low-resolution inputs are MODIS MOD13Q1 \SI{16}{day} composite time series at \SI{250}{m}. Each pixel carries a full annual trajectory of \(23\) composites (\(23\times\SI{16}{days}\), spanning the entire growing season) with six bands per date (NDVI, EVI, Red, NIR, Blue, MIR), giving the \(138\)-channel stacked temporal-spectral configuration used in training. We keep the composites in their native temporal order rather than reducing them to seasonal summaries, because the ordering itself carries the phenological cues (green-up timing, peak, senescence) that separate crops with similar peak greenness. Fig.~\ref{fig:modis_signatures} shows mean per-class signatures; at \SI{250}{m} the between-class differences are small and cereal trajectories overlap almost entirely, so no single band or date is decisive and the model must integrate weak evidence across the whole sequence.

\subsection{Class Mapping and Spatial Block Splits}
Raw AAFC labels are remapped to four crop classes: Spring Wheat, Barley, Canola, and Lentils, which are the most common crops in the study patches. Across the 751 patches, these four crops cover 50.8\% of all HR label pixels; within this four-crop pool, Canola accounts for 34.4\%, Lentils for 27.5\%, Spring Wheat for 27.1\%, and Barley for 11.1\%. The remaining pixels fall into two groups, both excluded from training and evaluation: non-crop land cover inside the patch footprints (roads, water, built-up areas, and other background; 46.6\%), and other AAFC classes such as Oats, Mustard, and Grassland (2.5\%). Excluding the non-crop background, the four target crops therefore cover about 95\% of the remaining pixels (50.8\% of 53.3\%), so restricting the task to these crops keeps most of the crop area while keeping the number of endmembers small.

The benchmark is therefore a \emph{closed-set} four-crop SPM task \emph{inside a known agricultural mask}: non-crop and minor-crop pixels are excluded from training and evaluation, and the reported metrics assume that such a mask is available at inference. We do not claim province-wide inventory mapping with automatic crop/non-crop separation; because the classifier is a softmax over only four classes, pixels outside the mask would otherwise be forced into one of the four crops. Patches with less than 10\% four-crop coverage are dropped when the data are loaded, and the 751 patches are those that remain after this filter.

\begin{figure}[!htbp]
\centering
\vspace{-4pt}
\includegraphics[width=\columnwidth]{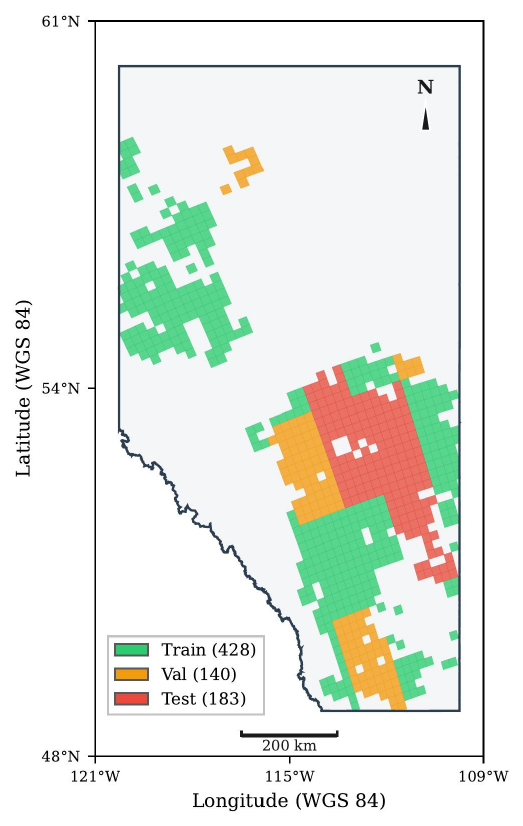}
\vspace{-4pt}
\caption{Spatial-block split for Alberta 2015. Each colored square is one patch, colored by its split: train (428 patches), validation (140), or test (183). Patches are assigned to splits by block of a \(6\times6\) grid, so each split forms contiguous areas. Patch footprints are clipped to the province outline (WGS~84). The same split is used for every model in Table~\ref{tab:main_results}.}
\label{fig:spatial_split}
\end{figure}

Train, validation, and test sets use a \emph{spatial block} split (Fig.~\ref{fig:spatial_split}). Alberta is divided into a \(6\times6\) grid of blocks, and every block is assigned as a whole to one split (seed 42):
\begin{itemize}
    \item Train: 428 patches
    \item Validation: 140 patches
    \item Test: 183 patches
\end{itemize}
Each patch covers \(64\times64\) LR pixels and \(512\times512\) HR pixels (\(\times 8\)), and the three sets sum to the full corpus (\(428{+}140{+}183{=}751\) patches; 57\%/19\%/24\%). A random patch split would place neighboring patches, which share atmosphere, phenology, and often the same physical fields, on both sides of the train/test boundary and inflate accuracy. Assigning whole blocks to one split prevents this: test blocks do not overlap training blocks, so the reported numbers measure transfer to unseen locations rather than memorization of neighbors. Compared with a conventional 70/15/15 random split, this assignment also keeps more patches for testing and fewer for training, which makes the benchmark stricter. The same split is used for every experiment in this article, and all quantitative results are computed over the 183 test patches.

\section{Methodology}
\label{sec:method}

\begin{figure*}[!tp]
\centering
\includegraphics[width=\textwidth]{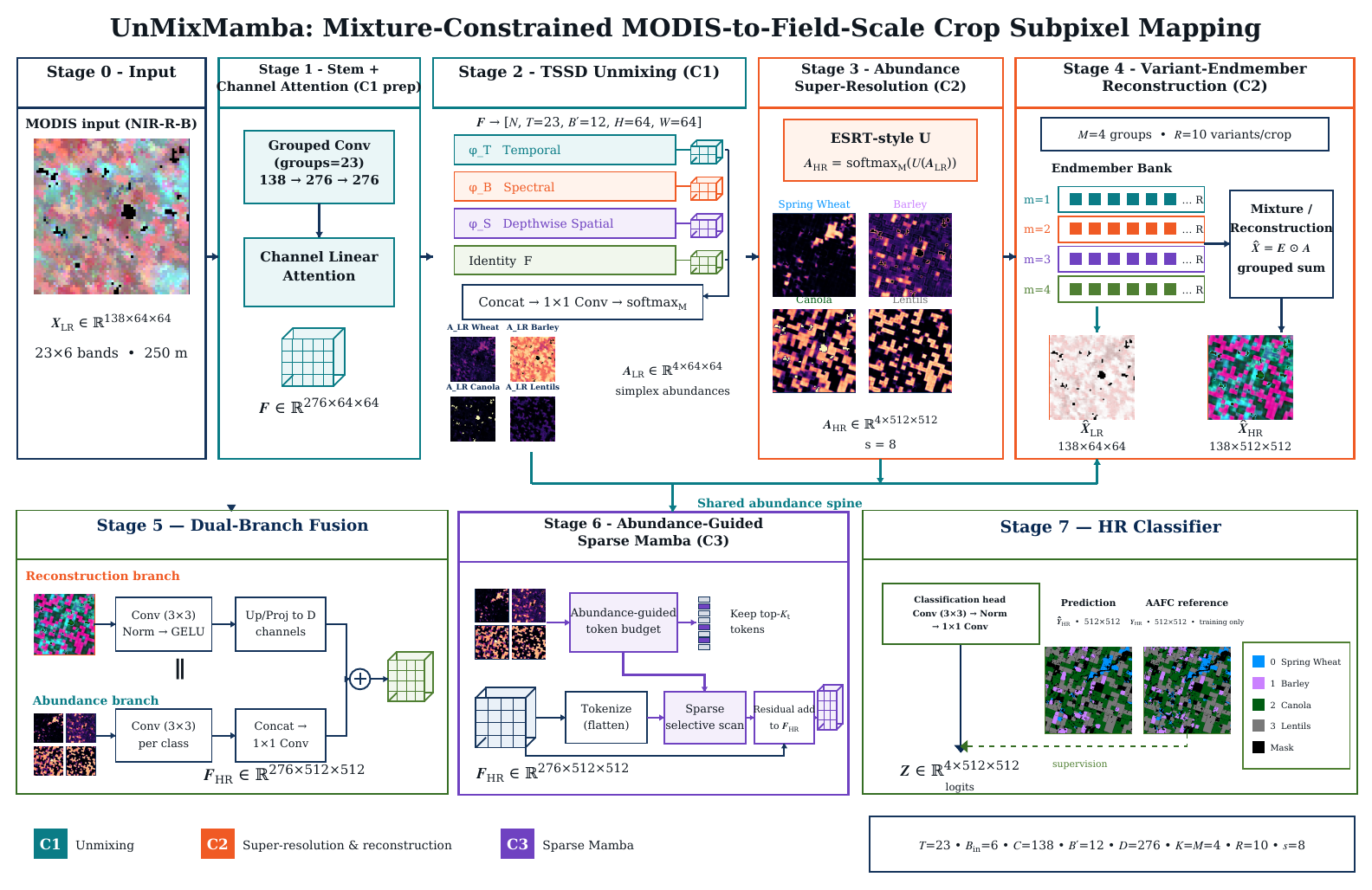}
\caption{UnMixMamba forward pipeline (Section~\ref{sec:arch}; Algorithm~\ref{alg:forward}). Top row: Stage~0, MODIS input \(X_{LR}\); Stage~1, grouped stem and channel linear attention; Stage~2, TSSD unmixing (C1), in which temporal, spectral, depthwise spatial, and identity branches are fused into the simplex abundances \(\mathbf{A}_{LR}\); Stage~3, abundance super-resolution to \(\mathbf{A}_{HR}\) (C2); and Stage~4, variant-endmember reconstruction of \(\hat{X}_{LR}\) and \(\hat{X}_{HR}\) (C2). Bottom row: Stage~5, dual-branch fusion of the reconstructed HR spectra and the HR abundances into \(F_{HR}\); Stage~6, abundance-guided sparse Mamba refinement (C3); and Stage~7, the HR classifier producing logits \(\mathbf{Z}\) and the hard map \(\hat{Y}_{HR}\), supervised by AAFC labels during training. The shared abundance spine carries \(\mathbf{A}_{LR}\) and \(\mathbf{A}_{HR}\) to the reconstruction, fusion, and token-budget stages. Example maps show an Alberta patch; the bottom strip lists the main hyperparameters (Table~\ref{tab:notation}).}
\label{fig:arch_flowchart}
\end{figure*}

\subsection{Problem Formulation and Notation}
Let \(X_{LR}\in\mathbb{R}^{C\times H\times W}\) denote an LR MODIS time-series patch with \(C{=}138\) channels organized as \(C{=}T\cdot B_{\mathrm{in}}\), where \(T{=}23\) sixteen-day composites and \(B_{\mathrm{in}}{=}6\) bands per composite (NDVI, EVI, Red, NIR, Blue, MIR). Let \(Y_{HR}\in\{0,\dots,K{-}1\}^{sH\times sW}\) be the co-registered AAFC label map at scale factor \(s{=}8\) (\(K{=}4\) crops; \(H{=}W{=}64\), giving HR size \(512{\times}512\)). UnMixMamba learns \(f_\theta: X_{LR}\!\mapsto\!\hat{Y}_{HR}\) with internal state organized around per-pixel abundance fields. Abundances are indexed by \(M\) endmember groups and logits by \(K\) classes; here each group corresponds to one crop, so \(M{=}K{=}4\), and each group owns \(R{=}10\) learnable spectral variants. Table~\ref{tab:notation} summarizes the notation; Table~\ref{tab:hyperparams} lists the hyperparameters of the model used for all reported results.

\begin{table}[!t]
\caption{Notation (see also Table~\ref{tab:hyperparams} and Algorithm~\ref{alg:forward}).}
\label{tab:notation}
\centering
\scriptsize
\setlength{\tabcolsep}{3pt}
\renewcommand{\arraystretch}{0.95}
\begin{tabular}{@{}ll@{}}
\toprule
\textbf{Symbol} & \textbf{Meaning} \\
\midrule
\(X_{LR}\) & LR MODIS patch, \(\mathbb{R}^{138\times 64\times 64}\) \\
\(T,\,B_{\mathrm{in}}\) & time steps (23); bands/step (6) \\
\(F\) & LR stem features, \(\mathbb{R}^{D\times H\times W}\) (\(D{=}276\)) \\
\(D,\,B'\) & stem width (276); TSSD bands/step (12) \\
\(s,\,K,\,M,\,R\) & upscale (\(\times 8\)); crops (4); groups (4); variants/group (10) \\
\(\mathbf{A}_{LR},\mathbf{A}_{HR}\) & LR/HR abundance maps (\(M\) ch.) \\
\(\mathbf{E},\,\mathbf{W}_{\mathrm{var}}\) & endmember bank; variant weights \\
\(\hat{X}_{LR},\hat{X}_{HR}\) & recon.\ cubes (full-channel LMM) \\
\(F_{HR},\,\mathbf{Z}\) & fused HR features; class logits (\(K\) ch.) \\
\(\lambda,\,\rho_u\) & token budget (0.1); unsel.\ ratio (0.3) \\
\(\parallel\) & parallel modules (shared input; Fig.~\ref{fig:arch_flowchart}) \\
\bottomrule
\end{tabular}
\end{table}

\begin{table}[!t]
\caption{Hyperparameters of UnMixMamba used for all reported results (Alberta 2015; full C1--C3 blocks).}
\label{tab:hyperparams}
\centering
\footnotesize
\setlength{\tabcolsep}{3pt}
\renewcommand{\arraystretch}{1.12}
\begin{minipage}[t]{0.48\columnwidth}
\centering
\textbf{(a) Architecture}\par\vspace{2pt}
\begin{tabular}{@{}>{\raggedright\arraybackslash}p{0.62\linewidth}r@{}}
\toprule
\textbf{Setting} & \textbf{Val.} \\
\midrule
Patch size (LR / HR) & $64^2$ / $512^2$ \\
Hidden dim.\ $D$ & 276 \\
Variants $R$ / crop & 10 \\
Token budget $\lambda$ & 0.1 \\
Mass / area weights & 0.7 / 0.3 \\
Abundance thresh. & 0.5 \\
Hard class boost & 0.03 \\
Unsel.\ ratio $\rho_u$ & 0.3 \\
\bottomrule
\end{tabular}
\end{minipage}\hfill
\begin{minipage}[t]{0.48\columnwidth}
\centering
\textbf{(b) Training}\par\vspace{2pt}
\begin{tabular}{@{}>{\raggedright\arraybackslash}p{0.62\linewidth}r@{}}
\toprule
\textbf{Setting} & \textbf{Val.} \\
\midrule
Epochs & 80 \\
Batch size & 2 \\
Optimizer / LR & Adam, $10^{-4}$ \\
EMA decay & 0.999 \\
\multicolumn{2}{@{}l@{}}{\footnotesize Hard classes: Barley, Canola.} \\
\multicolumn{2}{@{}l@{}}{\footnotesize $\rho_u$ cap: 10\,000 tokens.} \\
\bottomrule
\end{tabular}
\end{minipage}
\end{table}

\subsection{Architecture Overview}
\label{sec:arch}
Section~\ref{sec:dataset} and Table~\ref{tab:notation} define the input geometry; the forward path below is summarized in Algorithm~\ref{alg:forward}. The building blocks map to the contributions as follows: C1 realizes Contribution~1 (TSSD abundance SPM), C2 realizes Contribution~2 (variant-endmember LR/HR reconstruction and abundance SR), and C3 realizes Contribution~3 (abundance-guided sparse HR refinement).

\emph{Stage 0 (input).} \(X_{LR}\in\mathbb{R}^{138\times 64\times 64}\).

\emph{Stage 1: stem and LR encoding (C1 prep).} A timestep grouped convolutional stem maps \(X_{LR}\) to \(F\in\mathbb{R}^{276\times 64\times 64}\), followed by channel-wise linear attention.

\emph{Stage 2: TSSD unmixing (C1).} The Temporal Spectral Spatial Decoupler reshapes \(F\) as \([T,B',H,W]\) with \(T{=}23\), \(B'{=}12\) (\(TB'{=}D\)) and applies temporal, spectral, and depthwise spatial convolutions. The four branch tensors are fused and passed through a \(1{\times}1\) head to produce \(\mathbf{A}_{LR}=\mathrm{softmax}_M(\cdot)\in\mathbb{R}^{4\times 64\times 64}\).

\emph{Stage 3: abundance super resolution (C2, part 1).} An efficient super resolution transformer (ESRT)-style upsampler \(\mathcal{U}\) maps \(\mathbf{A}_{LR}\to \mathbf{A}_{HR}\in\mathbb{R}^{4\times 512\times 512}\) (softmax normalized).

\emph{Stage 4: variant endmember reconstruction (C2, part 2).} A learnable bank \(\mathbf{E}\in\mathbb{R}^{4\times 10\times 138}\) holds variant spectra. Per-pixel variant weights \(\mathbf{W}_{\mathrm{var}}\) are predicted at LR and HR and combined with abundances in Eq.~\eqref{eq:recon} to yield \(\hat{X}_{LR}\) and \(\hat{X}_{HR}\) (linear mix over all 138 channels, including NDVI/EVI as an approximation).

\emph{Stage 5: dual branch HR fusion.} \(\hat{X}_{HR}\) is patch-embedded (\(1{\times}1\) conv, grouped by timestep) and spatially attended; \(\mathbf{A}_{HR}\) is projected to \(D\) channels and channel-attended. The two \(D\)-channel maps are concatenated and fused by a \(3{\times}3\) conv to form \(F_{HR}\). The fusion is dual branch by design: one branch reads the reconstructed HR spectra (what the pixel looks like), the other reads the projected HR abundances (what the pixel is made of), so the downstream classifier can fall back on composition when appearance is ambiguous.

\emph{Stage 6: abundance-guided sparse Mamba (C3).} Token counts \(\{n_m\}\) are computed from \(\mathbf{A}_{HR}\) (Eqs.~\eqref{eq:token}--\eqref{eq:ntoken}). The top-\(n_m\) locations per crop enter crop-specific Mamba blocks; a random 30\% subset of remaining tokens uses a lighter Mamba path. Outputs are scattered back with a residual connection to \(F_{HR}\).

\emph{Stage 7: HR classification.} A two-layer conv head outputs \(\mathbf{Z}\in\mathbb{R}^{4\times 512\times 512}\). The hard map is \(\hat{Y}_{HR}=\arg\max_K \mathbf{Z}\).

UnMixMamba predicts \(\hat{Y}_{HR}\) through these seven stages (Algorithm~\ref{alg:forward}). Fig.~\ref{fig:arch_flowchart} diagrams the full forward path and the shared abundance spine across blocks C1--C3. The TSSD branches in Stage~2 and the two fusion branches in Stage~5 (marked \(\parallel\)) run in parallel on a shared input and are fused before the next stage.

\subsection{C1: Temporal Spectral Spatial Decoupled Unmixing}
\label{sec:c1}

Block C1 (Contribution~1) maps \(X_{LR}\) to simplex abundances. A grouped stem, with convolutions grouped by the \(T{=}23\) time steps, first maps \(X_{LR}\) to features \(F\in\mathbb{R}^{D\times H\times W}\) with \(D{=}276\). The TSSD module then reshapes \(F\) to \(\mathbb{R}^{T\times B'\times H\times W}\), with \(B'{=}12\) feature bands per time step (\(TB'{=}D\)), and applies three complementary operators \emph{in parallel}, together with an identity branch that keeps \(F\) (Fig.~\ref{fig:arch_flowchart}, Stage~2):

\begin{align}
F_{\mathrm{tmp}} &= \phi_{T}(F), \quad \text{(conv1d along } T\text{)} \\
F_{\mathrm{spe}} &= \phi_{B}(F), \quad \text{(conv1d along } B'\text{)} \\
F_{\mathrm{spa}} &= \phi_{S}(F), \quad \text{(depthwise spatial conv)} ,
\end{align}
where \(\phi_T\) acts along the temporal axis for each spatial location and intra-step band, \(\phi_B\) along the spectral axis for each timestep, and \(\phi_S\) is a depthwise spatial convolution. The input and three branches are concatenated, fused by a \(1\times1\) convolution, and passed to a \(1\times1\) abundance head; \(g(\cdot)\) denotes this fusion-plus-head mapping:
\begin{equation}
\mathbf{A}_{LR} = \mathrm{softmax}_M\!\big(\,g\big([\,F,\,F_{\mathrm{spa}},\,F_{\mathrm{tmp}},\,F_{\mathrm{spe}}\,]\big)\big)\in\mathbb{R}^{M\times H\times W}.
\end{equation}
The three axes of the MODIS cube carry different, non-interchangeable information: the temporal axis encodes phenological ordering, the intra-step spectral axis encodes instantaneous canopy state, and the spatial axis encodes field geometry. A generic convolution over all \(C{=}138\) flattened channels entangles these axes and lets a single spurious band-date interaction dominate; the decoupled branches instead let each axis be modeled by an operator suited to it and only fuse them at the abundance head. Emitting a softmax abundance field rather than class logits at this stage is how block C1 realizes Contribution~1: every downstream module reads the same normalized fractions, so reconstruction, sparse refinement, and classification remain tied to one mixture representation.

\subsection{C2: Abundance Super Resolution and Variant Endmember Reconstruction}

\begin{figure*}[!tp]
\centering
\vspace{-6pt}
\includegraphics[width=\textwidth]{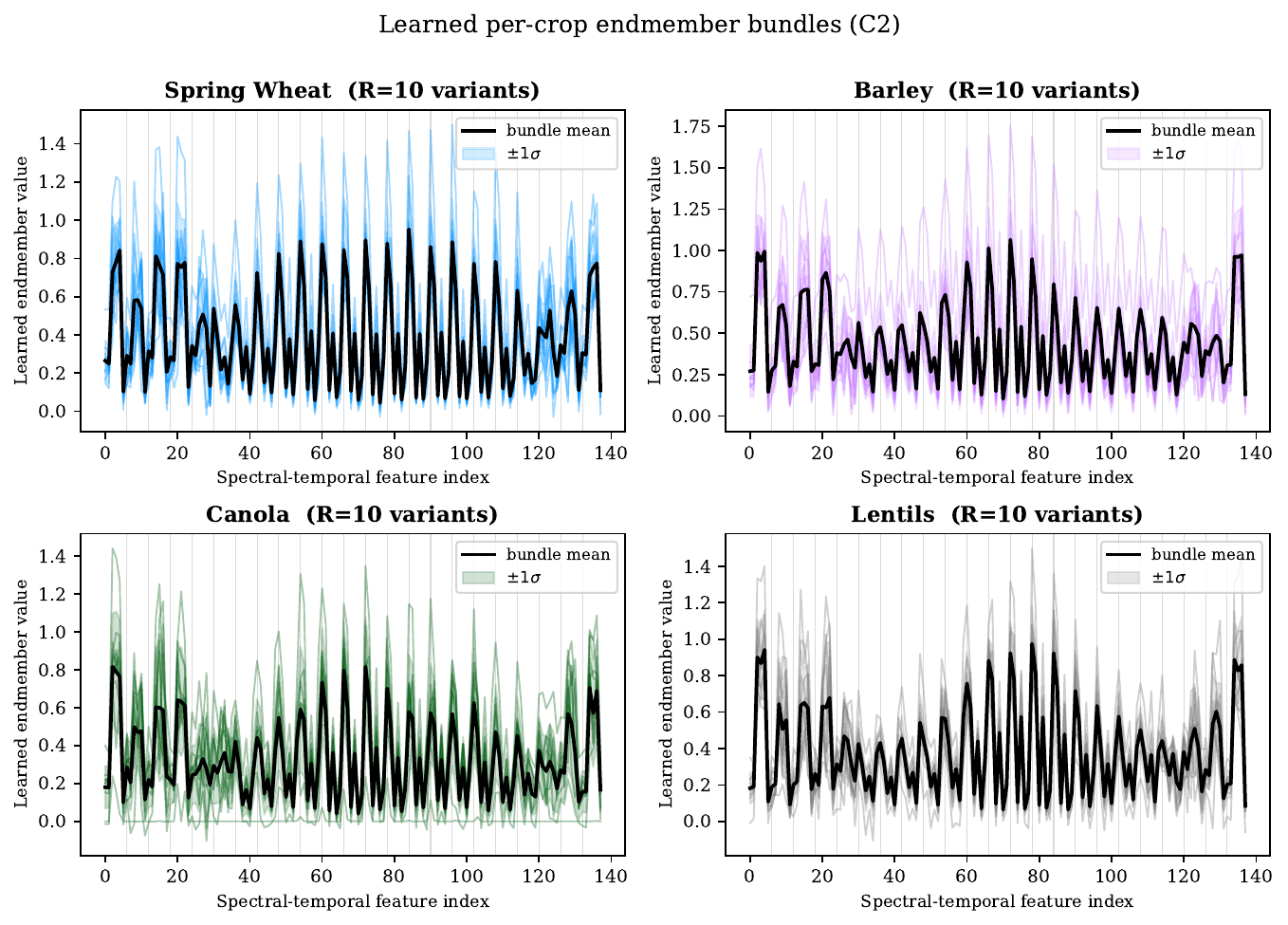}
\vspace{-4pt}
\caption{Learned endmember bundles (C2): $R{=}10$ variant spectra per crop (faint curves) with bundle mean and $\pm1\sigma$ envelope. Vertical ticks separate the 23 MODIS time steps ($\times$6 bands per step). Spring Wheat and Barley variants overlap phenologically, explaining cereal confusion; Canola and Lentils retain distinct temporal-spectral envelopes that the variant bank must represent separately.}
\label{fig:endmembers}
\end{figure*}

\begin{figure*}[!tp]
\centering
\vspace{-6pt}
\includegraphics[width=\textwidth]{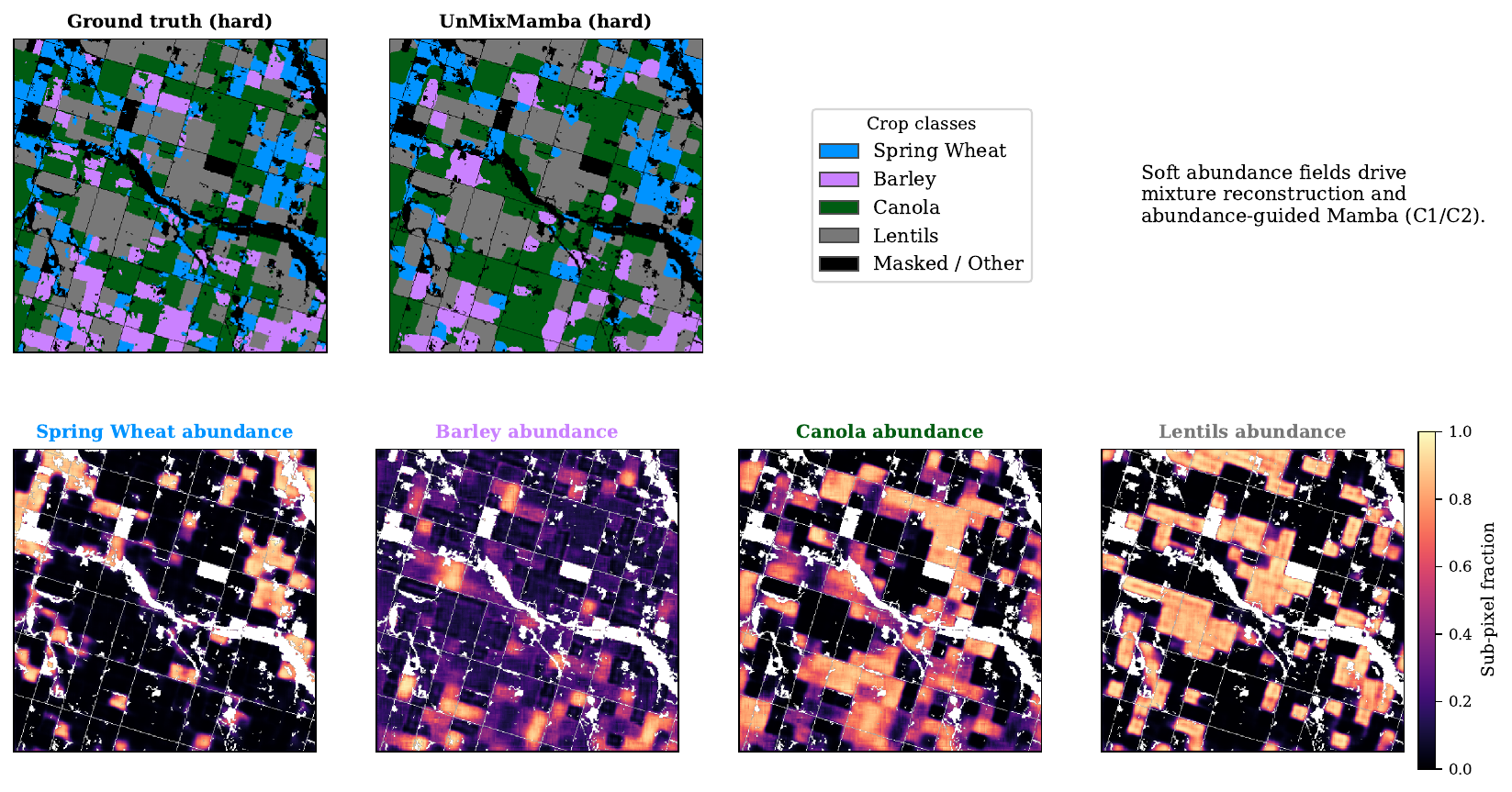}
\vspace{-4pt}
\caption{Soft HR abundance fields vs.\ hard prediction on a mixed test patch (C1/C2). Top row: AAFC ground truth and UnMixMamba hard map after argmax. Bottom row: per crop subpixel fraction maps predicted before classification. Bright fractional responses at field edges indicate compositional structure; soft abundances support reconstruction and C3 token budgeting.}
\label{fig:abundance}
\end{figure*}

Block C2 realizes Contribution~2 by super-resolving abundances and reconstructing spectra with learnable variants. LR abundances are super-resolved by an ESRT-style module \(\mathcal{U}\) to logits that are softmax normalized:
\begin{equation}
\mathbf{A}_{HR}=\mathrm{softmax}_M\!\big(\mathcal{U}(\mathbf{A}_{LR})\big)\in\mathbb{R}^{M\times sH\times sW}.
\end{equation}
Each group \(m\) owns \(R{=}10\) learnable spectral variants in \(\mathbf{E}\in\mathbb{R}^{M\times R\times C}\).
At each scale, per-pixel variant weights are predicted from abundance features and normalized over \(R\):
\begin{equation}
\mathbf{W}_{\mathrm{var}} = \mathrm{softmax}_R\!\big(h(\mathbf{A})\big),\qquad \textstyle\sum_{r} W_{\mathrm{var},m,r}=1 .
\end{equation}
Spectra are reconstructed under a variant aware linear mixture model over the full \(C{=}138\) channel stack:
\begin{equation}
\hat{X} \;=\; \sum_{m=1}^{M}\sum_{r=1}^{R} \mathbf{A}_{m}\; W_{\mathrm{var},m,r}\; \mathbf{E}_{m,r},
\label{eq:recon}
\end{equation}
applied separately at LR and HR to yield \(\hat{X}_{LR}\) and \(\hat{X}_{HR}\). Eq.~\eqref{eq:recon} is applied to all 138 channels, including NDVI and EVI. Because these indices are band ratios, they do not mix linearly with subpixel fractions, so their reconstruction under a linear mixing model (LMM) is an approximation that we retain for simplicity. Eq.~\eqref{eq:recon} nevertheless ties subpixel labels to an explicit mixture with \emph{intra-class spectral variability} rather than a single spectrum per crop. Super-resolving in the abundance domain rather than the radiance domain keeps the fine-scale quantity on the simplex (non-negative, sums to one per pixel), which is a far more constrained target than arbitrary HR radiance. The variant bundle \(\mathbf{E}\) gives each crop \(R{=}10\) spectra instead of one, so intra-class spectral diversity can be absorbed inside the crop rather than forcing a wrong-class decision. We treat these variants as learnable intra-class spectral degrees of freedom rather than as physically labeled factors; their utility is established functionally by ablation (Section~\ref{sec:ablation}). The reconstruction losses at both scales supervise the variants without any spectral ground truth beyond the MODIS input itself; the HR reconstruction term is a pooled consistency regularizer (Section~\ref{sec:loss}), not HR spectral supervision. Figs.~\ref{fig:endmembers} and~\ref{fig:abundance} visualize the learned variant bundles and the soft HR abundance fields that this stage produces.

\subsection{C3: Abundance Guided Sparse Mamba Refinement}
\label{sec:c3}

Block C3 implements Contribution~3. Dual-branch features \(F_{HR}\in\mathbb{R}^{D\times sH\times sW}\) are built as in Stage~5 above. Token budgets are computed from \(\mathbf{A}_{HR}\). For group \(m\), normalized abundance mass and thresholded area are combined as
\begin{equation}
\mathrm{score}_m =
0.7\,\widehat{\mathrm{mass}}_m +
0.3\,\widehat{\mathrm{area}}_m,
\label{eq:token}
\end{equation}
where area is computed from pixels with abundance above
\(0.5\,\max_m \mathbf{A}_{HR}\), and hats denote normalization across the \(M\) groups. The \(0.7/0.3\) weighting is a fixed design choice that places slightly greater emphasis on abundance mass than spatial extent. Sensitivity experiments showed negligible changes in accuracy for alternative weightings, indicating that performance is not strongly dependent on the exact values.

The per group token count is
\begin{equation}
n_m=\left\lceil \lambda\, sH\, sW\, \mathrm{score}_m\right\rceil,\quad \lambda=0.1,
\label{eq:ntoken}
\end{equation}
clamped to at least one token when mass is positive, with an additional floor of \(\lceil 0.03\, sH\, sW\rceil\) for Barley and Canola (indices 1--2). The top-\(n_m\) HR locations per group enter crop-specific Mamba blocks; a random subset (\(\rho_u{=}0.3\), capped at 10\,000 tokens) of unselected locations uses a lighter Mamba path. Outputs are scattered back and added residually to \(F_{HR}\). The mass/area weighting (\(0.7/0.3\)) favors total fraction mass while still using spatial extent, so a crop concentrated in a few near-pure pixels is not starved by a competitor that covers many partially occupied pixels. The hard-class floor applies to Barley and Canola (indices 1--2): Barley is the minority class and carries most residual error, and Canola, although the most prevalent class (Section~\ref{sec:dataset}), is the class into which much of the Barley leakage flows (Section~\ref{sec:results}). The floor therefore guarantees refinement capacity on both sides of the oilseed--cereal arm, while the Spring Wheat--Barley arm is handled by the margin loss (Section~\ref{sec:loss}).

Top-\(n_m\) selection based on crop abundance mainly chooses high-abundance, nearly pure pixels rather than mixed pixels. To target mixed or boundary pixels, other criteria such as entropy, margin, or boundary distance would be required. Therefore, with a fixed budget \(\lambda\), C3 focuses sparse refinement on minority-class crops and high-abundance locations, not on mixed or boundary regions. In practice, only about \(5.6\%\) of the selected tokens lie on AAFC field boundaries (Section~\ref{sec:complexity}).

The selected tokens are processed by Mamba in descending order of abundance rather than spatial order. Spatial information is preserved because the Mamba output is added back to \(F_{HR}\), which retains the original spatial layout.

Applying Mamba only to the selected tokens, instead of the full \(sH \times sW = 262{,}144\)-token high-resolution grid, keeps computation manageable at \(\lambda = 0.1\). During inference, unselected locations are deterministically subsampled by their flat pixel index before being processed by a lighter Mamba branch.

Abundance-guided token selection focuses on \emph{where} the token budget is allocated, rather than on whether Mamba is superior to other efficient linear-time sequence models. The same token allocation strategy could be used with alternative backbones, such as sparse linear-time Transformers.

\subsection{Training Objective}
\label{sec:loss}
The total loss combines class-balanced classification, cross-scale reconstruction, mixture regularizers, a classification--abundance consistency term, and a pairwise logit-margin term:
\begin{equation}
\mathcal{L} = \mathcal{L}_{\mathrm{cls}} + \mathcal{L}_{\mathrm{rec}} + \mathcal{L}_{\mathrm{reg}} + \mathcal{L}_{\mathrm{cons}} + \mathcal{L}_{\mathrm{margin}},
\label{eq:total_loss}
\end{equation}
where \(\mathcal{L}_{\mathrm{cls}}\) is class-balanced cross entropy (CE); \(\mathcal{L}_{\mathrm{rec}}=\lVert \hat{X}_{LR}-X_{LR}\rVert_1+\lVert \mathrm{Pool}_{s}(\hat{X}_{HR})-X_{LR}\rVert_1\) with \(\hat{X}\) from Eq.~\eqref{eq:recon}. The first term is the LR reconstruction constraint (the reconstructed LR cube must match the observed MODIS cube). The second term is a cross-scale \emph{consistency regularizer}: \(\mathrm{Pool}_{s}\) is average pooling by the upscale factor \(s{=}8\), used because no native HR MODIS spectra exist at field scale. Average pooling is used as a simple approximation of the MODIS point spread function (PSF). Although the true MODIS PSF is better represented by a Gaussian- or sinc-shaped kernel, average pooling preserves the mean signal and provides a practical approximation. All results reported in this study use average pooling rather than a detailed sensor model. This term constrains only the average value of each \(8\times8\) block and does not require the HR reconstruction to be spatially uniform. A \SI{250}{m} pixel that covers a field boundary can still contain detailed spatial variation, provided that the average of the corresponding HR pixels matches the coarse observation. Next, \(\mathcal{L}_{\mathrm{reg}}=0.2\,\mathcal{L}_{\mathrm{dice}}+0.05\,\mathcal{L}_{\mathrm{SAM}}+0.0005\,\mathcal{L}_{\mathrm{TV}}\) on abundances and reconstructions; \(\mathcal{L}_{\mathrm{cons}}\) (weight 0.05) matches downsampled HR logits to \(\mathbf{A}_{LR}\); and \(\mathcal{L}_{\mathrm{margin}}\) adds weighted Spring Wheat--Barley (0.15) and Barley--Canola (0.08) logit-margin terms. \(\mathcal{L}_{\mathrm{cls}}\) and \(\mathcal{L}_{\mathrm{rec}}\) have unit weight.

\(\mathcal{L}_{\mathrm{cls}}\) is the main label-supervised term and drives the final decision; class balancing counteracts the Barley minority. \(\mathcal{L}_{\mathrm{margin}}\) also uses the AAFC labels to identify the true class of each pixel, whereas \(\mathcal{L}_{\mathrm{rec}}\) is supervised only by the MODIS input. \(\mathcal{L}_{\mathrm{rec}}\) is the mixture anchor shared by Contributions~1 and~2: because it forces the abundances and variant endmembers to regenerate the observed LR cube (and to stay consistent when the HR field is pooled back), it prevents the network from producing abundances that classify well but violate the mixing model. Within \(\mathcal{L}_{\mathrm{reg}}\), the Dice term keeps minority abundances from vanishing, the spectral angle (SAM) term makes reconstruction insensitive to illumination scaling, and the total variation (TV) term suppresses salt and pepper artifacts in the HR abundance field. \(\mathcal{L}_{\mathrm{cons}}\) enforces scale consistency by requiring the HR decision, when pooled back to \SI{250}{m}, to agree with the LR abundance field, closing the loop between the coarse and fine representations. \(\mathcal{L}_{\mathrm{margin}}\) pushes decision boundaries apart for both arms of the three-way Barley hub. Greater emphasis is placed on the Spring Wheat--Barley pair because cereal--cereal confusion is more challenging, while a smaller emphasis is applied to the Barley--Canola pair, which retains better mid-season spectral separation (Fig.~\ref{fig:modis_signatures}). The weights are fixed across all experiments; full optimizer settings are provided in Section~\ref{sec:impl}.

\begin{algorithm}[!t]
\caption{UnMixMamba forward pass}
\label{alg:forward}
\begin{algorithmic}[1]
\Require \(X_{LR}\in\mathbb{R}^{138\times 64\times 64}\)
\State \(F \gets \mathrm{Stem}(X_{LR})\) \hfill \(\triangleright\) \(F\in\mathbb{R}^{276\times 64\times 64}\)
\State \(F \gets \mathrm{ChannelAttn}(F)\)
\State \(\mathbf{A}_{LR}\gets \mathrm{softmax}_M(\mathrm{TSSD}(F))\) \Comment{C1; \(4\times 64\times 64\)}
\State \(\mathbf{A}_{HR}\gets \mathrm{softmax}_M(\mathcal{U}(\mathbf{A}_{LR}))\) \Comment{\(4\times 512\times 512\)}
\State \(\hat{X}_{LR},\hat{X}_{HR}\gets \mathrm{Mix}(\mathbf{A},\mathbf{W}_{\mathrm{var}},\mathbf{E})\) \Comment{C2, Eq.~\eqref{eq:recon}}
\State \(F_{HR}\gets \mathrm{Fuse}(\mathrm{Embed}(\hat{X}_{HR}),\,\mathrm{Proj}(\mathbf{A}_{HR}))\) \Comment{\(276\times 512\times 512\)}
\State \(\{n_m\}\gets \mathrm{TokenBudget}(\mathbf{A}_{HR};\,\lambda{=}0.1,\,0.7/0.3)\) \Comment{C3}
\State \(F_{HR}\gets F_{HR}+\mathrm{SparseMamba}(F_{HR},\{n_m\};\,\rho_u{=}0.3)\)
\State \(\mathbf{Z}\gets \mathrm{ClsHead}(F_{HR})\)
\State \Return \(\hat{Y}_{HR}=\arg\max_K \mathbf{Z}\)
\end{algorithmic}
\end{algorithm}

\section{Experimental Setup}
\label{sec:exp}

\subsection{Fair Protocol}
\label{sec:fair_protocol}
A recurring problem in SPM comparisons is that gains attributed to a model can instead come from test time tricks or from unequal training budgets. We therefore train and evaluate every method under one shared fair protocol. All methods use the same spatial block train/validation/test patches (Section~\ref{sec:dataset}), the same four-crop remapping and evaluation mask, and the same inference settings: no test-time augmentation and no median post-filtering. Every neural network in Table~\ref{tab:main_results} (UnMixMamba, ST-DRes, ESPCN, FineCrop, AgriFM, UNet, LSTM, and Transformer) is trained with the same 80 epoch budget, validation every 5 epochs with early stopping at patience 6, Adam at learning rate $10^{-4}$, and class-balanced cross entropy as the classification loss; UnMixMamba additionally uses the auxiliary loss terms of Section~\ref{sec:loss}.
Reported checkpoints are the validation-OA-best weights within this budget; for the strongest baseline, ST-DRes, the best validation OA occurs at the end of the budget.
The RF baseline uses the same train split, MODIS inputs, and AAFC labels but, as a non-neural model, is trained without epochs or a gradient optimizer. The classical SPM methods allocate labels from fractions and require no epoch-based training; following the standard two stage SPM pipeline, each classical allocator receives per-class abundance maps estimated at LR by a random forest regressor trained on the spatial block train split (identical MODIS inputs and AAFC labels), then allocates HR labels on the same 183 test patches.
Under this shared fair protocol, differences in Table~\ref{tab:main_results} reflect the mapping models themselves.

\subsection{Compared Methods}
\label{sec:baselines}
We compare UnMixMamba with thirteen baselines on the Alberta 2015 test split (Table~\ref{tab:main_results}): five classical SPM allocators; six deep or statistical mapping baselines (ST-DRes, ESPCN, UNet, LSTM, Transformer, and RF); and two crop mapping architectures (FineCrop and AgriFM) adapted to the MODIS$\times 8$ setting.

\emph{(i) Classical subpixel mapping methods.} Five classical algorithms were re-implemented from reference MATLAB/Python code and operate on LR abundance maps supplied by the RF regressor described in Section~\ref{sec:fair_protocol}: the subpixel/pixel spatial attraction model (SAPS) of Mertens \emph{et al.} \cite{mertens2006saps}; the genetic algorithm attraction variant (GAAI) \cite{mertens2003gaai}; radial basis function interpolation SPM (RBF) of Wang \emph{et al.} \cite{wang2014rbf}; the simulated annealing allocation of Ling \emph{et al.} (Ling) \cite{ling2010sa}; and a spectral spatial pixel swapping variant (SPM-SS) built on the spatial-dependence maximization principle of Atkinson \cite{atkinson2005psa}. SAPS, RBF, and pixel swapping are standard references in the SPM literature \cite{mertens2006saps,wang2014rbf,atkinson2005psa}. RBF and SPM-SS share identical scores in Tables~\ref{tab:main_results} and~\ref{tab:cross_province_sask} because, under the shared RF abundance input on this dataset, both allocators converge to the same HR label field.

\emph{(ii) Deep and statistical mapping baselines.} ST-DRes \cite{stdres2022} is the spatial-temporal depth-wise residual network of Wang \emph{et al.} and the dedicated MODIS crop SPM baseline; we use the authors' implementation (PixelShuffle upsampling, default depth configuration) retrained under the fair protocol. UNet is the bilinear upsampled encoder--decoder baseline from the same study \cite{stdres2022}. ESPCN \cite{espcn2016}, LSTM \cite{lstm1997}, Transformer \cite{transformer2017}, and RF \cite{breiman2001rf} are general-purpose models trained as LR$\rightarrow$HR mapping backbones with identical inputs, labels, and splits. The Swin Transformer baseline reported in \cite{stdres2022} is omitted because it is not part of the authors' released SPM codebase and lies outside our compute budget for a full retrain. Several related crop mapping models listed in Table~\ref{tab:novelty_audit} were also not included as quantitative baselines because no public, drop in implementation for MODIS to AAFC $\times 8$ SPM was available: PHAT \cite{phat2025}, CropSTS \cite{cropsts2025}, the EnMAP dual stream model \cite{enmap2025}, and TAS2B-Net \cite{tas2b2025} do not release usable training code for this setting (CropSTS reports PASTIS results, but the PASTIS benchmark repository provides the dataset/evaluation harness rather than the CropSTS model).

\emph{(iii) Crop-mapping architectures adapted to SPM.} FineCrop \cite{finecrop2025} and AgriFM \cite{agrifm2025} are adapted to the MODIS$\times 8$ task and trained and evaluated under the same fair protocol as the other learned baselines.

\subsection{Metrics}
We report overall accuracy (OA, the fraction of correctly mapped HR pixels), mean intersection over union (mIoU, the mean of per-class IoU), average accuracy (AA, the mean of per-class producer accuracies), the Cohen's Kappa coefficient, and per-class producer accuracy (equivalently, recall; the per-class columns of all result tables). OA measures global correctness, AA emphasizes performance on minority classes by weighting all classes equally, and mIoU provides a stricter assessment by jointly accounting for omission and commission errors. mIoU therefore complements OA and AA, while Kappa measures agreement beyond chance. All metrics are computed at pixel level and aggregated across all valid HR pixels of the 183 held-out test patches (masked/other pixels excluded).

\subsection{Implementation Details}
\label{sec:impl}
UnMixMamba uses hidden dimension \(D{=}276\), \(M{=}4\) crop groups, and \(R{=}10\) endmember variants per group (unless ablated). Input channels are \(C{=}138\) (\(T{=}23\) composites, \(B_{\mathrm{in}}{=}6\) bands each); TSSD operates on stem features with \(B'{=}12\) bands per composite (\(TB'{=}D\)). Training uses Adam (\(10^{-4}\)), class-balanced cross entropy, mixed precision, and an exponential moving average (EMA) of weights (decay \(0.999\)) for 80 epochs (batch size 2); the remaining loss terms follow Section~\ref{sec:loss}. Hard class token indices are Barley and Canola; \(\lambda{=}0.1\), hard class boost \(=0.03\), and unselected token ratio \(\rho_u{=}0.3\).
All other training and evaluation settings follow the fair protocol (Section~\ref{sec:fair_protocol}).

\section{Results and Discussion}
\label{sec:results}

Table~\ref{tab:main_results} reports the primary benchmark comparison against thirteen baselines on the held-out Alberta 2015 test split (183 patches); Section~\ref{sec:qualitative} gives visual comparisons. Sections~\ref{sec:ablation}--\ref{sec:complexity} analyze UnMixMamba internally (block ablations, learned mixtures, multi-seed retraining, token budget sweeps, and computational cost), and Section~\ref{sec:cross_province} tests geographic transfer to Saskatchewan. We interpret results against the three contributions in Section~\ref{sec:contributions}.

\subsection{Quantitative Comparison with Baselines}
\label{sec:quantitative}
Table~\ref{tab:main_results} summarizes test-set performance on Alberta 2015 under the fair protocol (183 test patches).

\begin{table*}[!tp]
\caption{Test-set accuracy on Alberta 2015 (four crops) under the same fair protocol (Section~\ref{sec:fair_protocol}). Per-class columns are producer accuracy (recall). Best value per column in bold; second best underlined.}
\label{tab:main_results}
\centering
\footnotesize
\setlength{\tabcolsep}{3pt}
\begin{tabular}{lccccccccc}
\toprule
\textbf{Model} & \textbf{OA} & \textbf{mIoU} & \textbf{AA} & \textbf{Kappa} & \textbf{Spring Wheat} & \textbf{Barley} & \textbf{Canola} & \textbf{Lentils} \\
\midrule
\textbf{UnMixMamba} & \textbf{0.8397} & \textbf{0.6640} & \textbf{0.7837} & \textbf{0.7715} & \underline{0.8751} & \textbf{0.5496} & \underline{0.8370} & \textbf{0.8733} \\
ST-DRes & \underline{0.7970} & 0.5662 & 0.7060 & \underline{0.7100} & 0.8301 & \underline{0.4376} & 0.7514 & 0.8049 \\
FineCrop & 0.7943 & 0.5723 & 0.6893 & 0.7038 & 0.8667 & 0.2506 & 0.7988 & \underline{0.8411} \\
ESPCN & 0.7833 & \underline{0.5833} & \underline{0.7113} & 0.7030 & 0.8503 & 0.3738 & 0.7980 & 0.8234 \\
UNet & 0.7774 & 0.5560 & 0.6383 & 0.6881 & \textbf{0.8753} & 0.0000 & \textbf{0.8476} & 0.8305 \\
Transformer & 0.7694 & 0.5694 & 0.7056 & 0.6834 & 0.8479 & 0.4112 & 0.7788 & 0.7844 \\
AgriFM & 0.7694 & 0.5404 & 0.6654 & 0.6695 & 0.8549 & 0.2236 & 0.7423 & 0.8409 \\
LSTM & 0.7678 & 0.5632 & 0.6959 & 0.6803 & 0.8520 & 0.3653 & 0.7836 & 0.7826 \\
GAAI & 0.7462 & 0.5241 & 0.6434 & 0.6456 & 0.8491 & 0.1786 & 0.7924 & 0.7535 \\
SAPS & 0.7461 & 0.5241 & 0.6434 & 0.6456 & 0.8490 & 0.1787 & 0.7923 & 0.7535 \\
Ling & 0.7459 & 0.5238 & 0.6432 & 0.6452 & 0.8487 & 0.1790 & 0.7919 & 0.7532 \\
RF & 0.7122 & 0.4855 & 0.6125 & 0.5966 & 0.8265 & 0.1676 & 0.7593 & 0.6963 \\
RBF & 0.6143 & 0.3984 & 0.5515 & 0.4697 & 0.7072 & 0.2607 & 0.6130 & 0.6252 \\
SPM-SS & 0.6143 & 0.3984 & 0.5515 & 0.4697 & 0.7072 & 0.2607 & 0.6130 & 0.6252 \\
\bottomrule
\end{tabular}
\end{table*}

UnMixMamba reaches OA 0.8397, mIoU 0.6640, AA 0.7837, and Kappa 0.7715, ahead of every classical SPM method, FineCrop, AgriFM, and the strongest deep SPM baseline, ST-DRes (OA 0.7970, mIoU 0.5662). A paired Wilcoxon signed rank test over all $n{=}183$ test patches (per patch OA), with Holm correction over two planned comparisons, against ST-DRes (best baseline OA and Kappa) and ESPCN (best baseline mIoU and AA), finds both comparisons significant at family-wise $\alpha{=}0.05$ (each unadjusted $p < 10^{-12}$). FineCrop reaches OA 0.7943 with weak Barley (0.2506); AgriFM reaches OA 0.7694 with Barley 0.2236. UnMixMamba uses an explicit mixture-constrained abundance field; ST-DRes is a strong end-to-end spatial-temporal mapper without an explicit simplex or reconstruction loss. Against ESPCN, UnMixMamba gains +5.6~pp OA, +8.1~pp mIoU, and +17.6~pp Barley recall (0.5496 vs.\ 0.3738). The gap versus ST-DRes is +4.3~pp OA and +9.8~pp mIoU, and UnMixMamba also leads under multi-seed retraining (mean OA 0.836 vs.\ 0.797; Table~\ref{tab:multiseed_summary}). Because OA is weighted by class prevalence (Canola, Lentils, and Spring Wheat together account for about 89\% of four-crop pixels), AA and mIoU, which weight classes equally, are the fairer summaries; UnMixMamba leads on both, as well as on Kappa. UnMixMamba attains the best Barley (0.5496) and Lentils (0.8733) recall, and its Spring Wheat recall (0.8751) is within 0.0002 of the best. UNet reaches the best Spring Wheat (0.8753) and Canola (0.8476) recall but collapses Barley to 0.0000.

\begin{figure*}[!tp]
\centering
\vspace{-6pt}
\includegraphics[width=\textwidth]{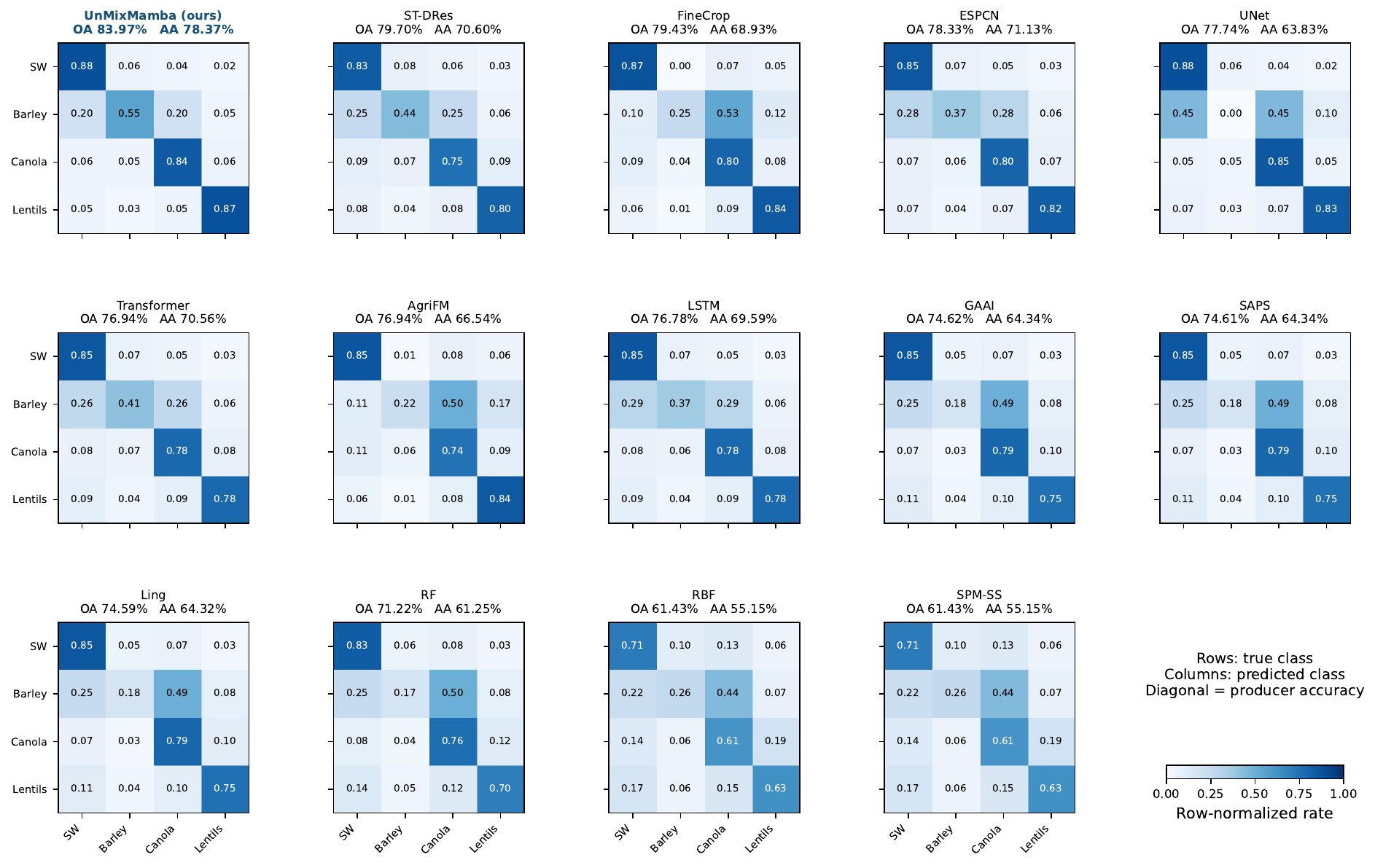}
\vspace{-4pt}
\caption{Normalized confusion matrices for all methods in Table~\ref{tab:main_results} on the held-out Alberta 2015 test split. Rows are true classes and columns are predicted classes; subtitles report OA and AA. UnMixMamba is highlighted in blue.}
\label{fig:confusion_all}
\end{figure*}

\begin{table*}[!tp]
\caption{Ablation and token-budget experiments for UnMixMamba on Alberta 2015 (183 test patches, 80 epoch fair protocol). Per-class columns are producer accuracy (recall); SW = Spring Wheat. Best value per column in bold; second best underlined. P0: UnMixMamba as in Table~\ref{tab:main_results}. A1--A3: primary C1--C3 ablations; A4--A6: C2/C3 refinements. T1--T2: token budget \(\lambda\) (the default \(\lambda{=}0.1\) is row P0). Multi-seed statistics: Table~\ref{tab:multiseed_summary}.}
\label{tab:all_unmix_experiments}
\centering
\scriptsize
\setlength{\tabcolsep}{2pt}
\renewcommand{\arraystretch}{1.05}
\begin{tabular}{cllccccccc}
\toprule
\textbf{ID} & \textbf{Experiment} & \textbf{Setting} & \textbf{OA} & \textbf{AA} & \textbf{Kap.} & \textbf{SW} & \textbf{Barley} & \textbf{Canola} & \textbf{Lentils} \\
\midrule
P0 & UnMixMamba (main) & $\lambda{=}0.1$ & \textbf{0.8397} & \textbf{0.7837} & \textbf{0.7715} & 0.8751 & \textbf{0.5496} & 0.8370 & \underline{0.8733} \\
\midrule
A1 & w/o C1 & shallow \(1{\times}1\) LR encoder & 0.6181 & 0.4890 & 0.4313 & 0.7167 & 0.0000 & \textbf{0.9376} & 0.3019 \\
A2 & w/o C2 & single variant query & 0.7778 & 0.6599 & 0.6769 & 0.8256 & 0.1521 & \underline{0.9182} & 0.7435 \\
A3 & w/o C3 & uniform token budget & 0.6418 & 0.5200 & 0.4780 & \textbf{0.9464} & 0.0000 & 0.7843 & 0.3491 \\
A4 & w/o C2 (bilinear) & bilinear abundance SR & 0.7738 & 0.6637 & 0.6739 & 0.8330 & 0.1983 & 0.7302 & \textbf{0.8932} \\
A5 & w/o C2 (no recon) & LR/HR recon.\ losses off & 0.7818 & 0.7143 & 0.6905 & 0.8613 & 0.4227 & 0.7558 & 0.8174 \\
A6 & w/o C3 (floor only) & uniform tokens + floor & 0.7537 & 0.6123 & 0.6414 & 0.8496 & 0.0239 & 0.7727 & 0.8028 \\
\midrule
T1 & $\lambda{=}0.05$ & sparse Mamba tokens & 0.7539 & 0.6164 & 0.6414 & 0.8005 & 0.0000 & 0.8803 & 0.7847 \\
T2 & $\lambda{=}0.50$ & higher-budget sparse (not dense) & \underline{0.8181} & \underline{0.7645} & \underline{0.7436} & \underline{0.8830} & \underline{0.5055}  & 0.8057 & 0.8637 \\
\bottomrule
\end{tabular}
\end{table*}

\begin{table*}[!tp]
\caption{Cost and accuracy of UnMixMamba and the deep SPM and general-purpose baselines on Alberta 2015 (single NVIDIA RTX A6000 GPU; $64\times64$ LR$\rightarrow$$512\times512$ HR). OA/mIoU from Table~\ref{tab:main_results}. GFLOPs estimated with THOP ($2\times$MACs) on a $1{\times}138{\times}64{\times}64$ input; UnMixMamba at default $\lambda{=}0.1$. Best accuracy in bold; second best underlined.}
\label{tab:complexity}
\centering
\footnotesize
\setlength{\tabcolsep}{6pt}
\begin{tabular}{lccccc}
\toprule
\textbf{Model} & \textbf{Params (M)} & \textbf{GFLOPs} & \textbf{Time (ms)} & \textbf{OA} & \textbf{mIoU} \\
\midrule
\textbf{UnMixMamba} ($\lambda{=}0.1$) & 6.92 & 1096 & 103.1 & \textbf{0.8397} & \textbf{0.6640} \\
ST-DRes & 0.05 & 0.4 & 1.0 & \underline{0.7970} & 0.5662 \\
ESPCN & 0.42 & 3.5 & 0.2 & 0.7833 & \underline{0.5833} \\
UNet & 2.48 & 51.5 & 2.2 & 0.7774 & 0.5560 \\
Transformer & 0.18 & 0.8 & 1.9 & 0.7694 & 0.5694 \\
LSTM & 0.14 & 1.8 & 30.7 & 0.7678 & 0.5632 \\
\bottomrule
\end{tabular}
\end{table*}

\begin{table*}[!tp]
\caption{Geographic transfer of Alberta trained methods to Saskatchewan 2015 (1{,}049 patches) under the same fair protocol. Rows follow the order of Table~\ref{tab:main_results}; per-class columns are producer accuracy (recall). Best value per column in bold; second best underlined.}
\label{tab:cross_province_sask}
\centering
\footnotesize
\setlength{\tabcolsep}{3pt}
\begin{tabular}{lccccccc}
\toprule
\textbf{Model} & \textbf{OA} & \textbf{AA} & \textbf{Kappa} & \textbf{Spring Wheat} & \textbf{Barley} & \textbf{Canola} & \textbf{Lentils} \\
\midrule
\textbf{UnMixMamba} & \textbf{0.7348} & \textbf{0.6841} & \textbf{0.6118} & 0.8360 & \textbf{0.4594} & 0.6953 & 0.7456 \\
ST-DRes & \underline{0.7253} & \underline{0.6671} & \underline{0.6015} & 0.8491 & \underline{0.3998} & 0.6614 & \textbf{0.7580} \\
FineCrop & 0.6730 & 0.6157 & 0.5310 & 0.8065 & 0.3324 & 0.5658 & \textbf{0.7580} \\
ESPCN & 0.7219 & 0.6346 & 0.5913 & 0.8399 & 0.2740 & 0.6753 & \underline{0.7493} \\
UNet & 0.7194 & 0.5716 & 0.5822 & \underline{0.8657} & 0.0000 & 0.7003 & 0.7206 \\
Transformer & 0.6883 & 0.6198 & 0.5470 & 0.8388 & 0.3106 & 0.6317 & 0.6983 \\
AgriFM & 0.6732 & 0.5923 & 0.5277 & \textbf{0.8685} & 0.2031 & 0.5570 & 0.7407 \\
LSTM & 0.6911 & 0.6161 & 0.5506 & 0.8348 & 0.2811 & 0.6206 & 0.7280 \\
GAAI & 0.7153 & 0.5644 & 0.5620 & 0.7007 & 0.0818 & \textbf{0.8260} & 0.6490 \\
SAPS & 0.7153 & 0.5644 & 0.5619 & 0.7007 & 0.0819 & \underline{0.8259} & 0.6490 \\
Ling & 0.7151 & 0.5643 & 0.5616 & 0.7005 & 0.0822 & 0.8256 & 0.6489 \\
RF & 0.6862 & 0.5419 & 0.5176 & 0.6779 & 0.0807 & 0.7983 & 0.6106 \\
RBF & 0.5175 & 0.4459 & 0.3019 & 0.5371 & 0.1947 & 0.5322 & 0.5195 \\
SPM-SS & 0.5175 & 0.4459 & 0.3019 & 0.5371 & 0.1947 & 0.5322 & 0.5195 \\
\bottomrule
\end{tabular}
\end{table*}

The confusion matrices in Fig.~\ref{fig:confusion_all} show a largely uniform error structure across methods: every model keeps Spring Wheat, Canola, and Lentils on the diagonal, and the remaining off-diagonal mass is concentrated in the Barley row, where true Barley pixels leak mainly into Spring Wheat and Canola. This is consistent with Fig.~\ref{fig:modis_signatures}: Barley shares its phenological envelope with Spring Wheat and its mid-season canopy with Canola, so at \SI{250}{m} it has no private region of feature space. The classical allocators yield Barley recall of \(0.18\)--\(0.26\); FineCrop and AgriFM reach 0.2506 and 0.2236, respectively; and the strongest general-purpose deep baselines recover \(0.37\)--\(0.44\). UnMixMamba attains the highest Barley recall (0.5496) while keeping Spring Wheat, Canola, and Lentils competitive, i.e., it redistributes error out of the Barley row without simply relabeling one cereal as the other.

\subsection{Qualitative Comparison}
\label{sec:qualitative}
\begin{figure*}[!tp]
\centering
\includegraphics[width=\textwidth,height=0.86\textheight,keepaspectratio]{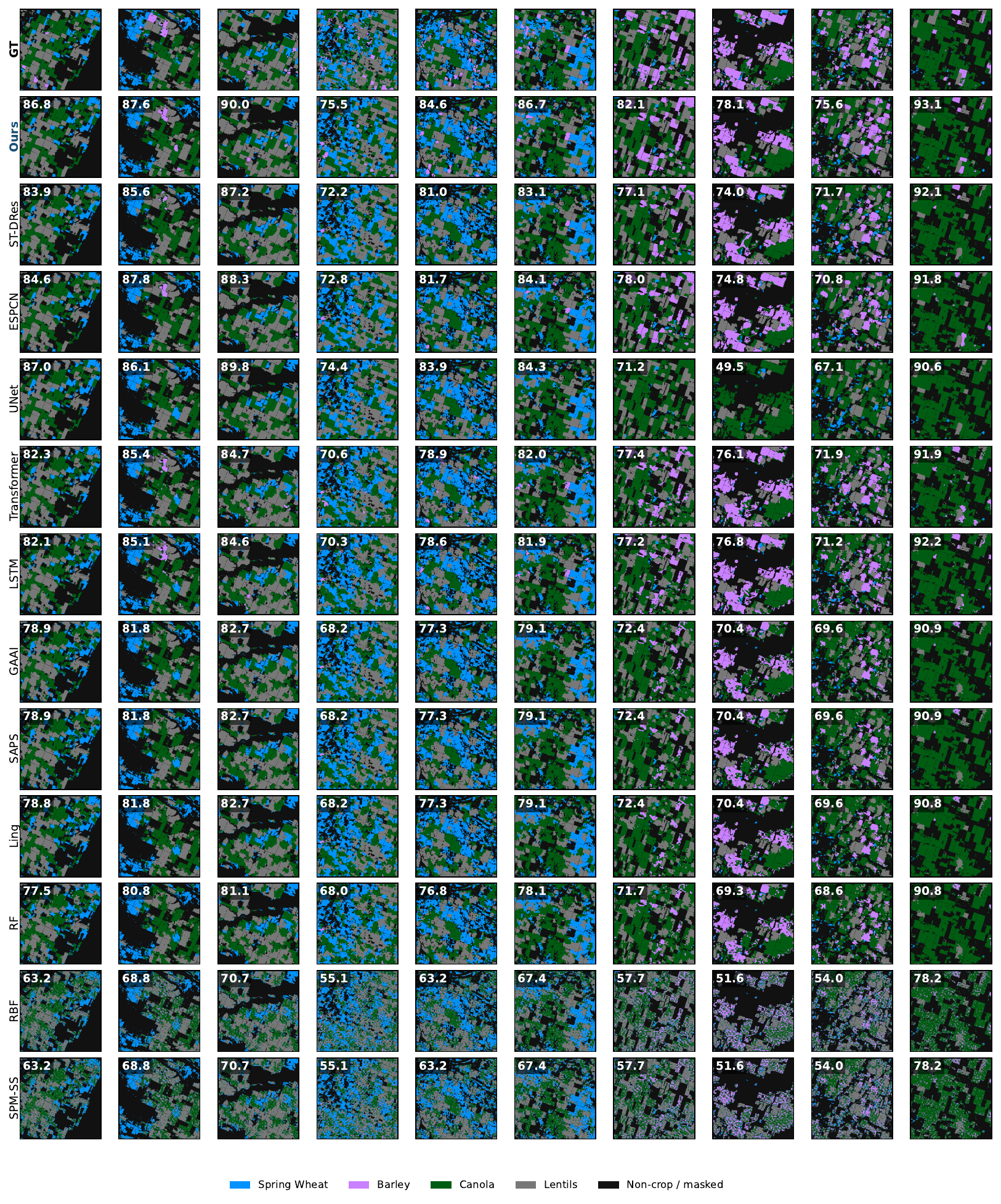}
\caption{Subpixel maps on ten randomly selected Alberta patches. Rows: ground truth (GT), UnMixMamba, and the baselines of Table~\ref{tab:main_results}; columns are patches. Upper-left numbers are panel OA (\%). Colors: Spring Wheat (blue), Barley (purple), Canola (green), Lentils (grey); masked pixels are black. Tiles: r22\_c32, r26\_c23, r28\_c25, r46\_c37, r53\_c50, r57\_c44, r73\_c32, r75\_c27, r77\_c31, r80\_c36.}
\label{fig:qualitative_random10}
\end{figure*}

\begin{figure*}[!tp]
\centering
\vspace{-6pt}
\includegraphics[width=\textwidth]{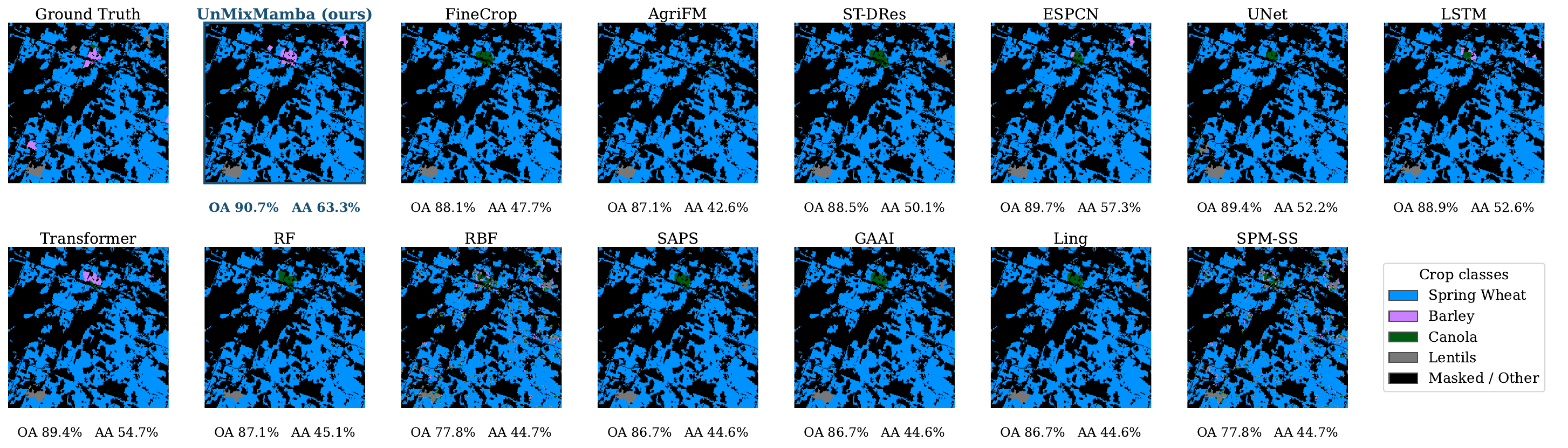}
\vspace{-4pt}
\caption{Spring Wheat-focused comparison on an Alberta patch. A $256\times256$ HR window is centered on Spring Wheat pixels. Subtitles report patch-level OA and AA; UnMixMamba is highlighted in blue.}
\label{fig:qual_focus_sw}
\end{figure*}

\begin{figure*}[!tp]
\centering
\vspace{-6pt}
\includegraphics[width=\textwidth]{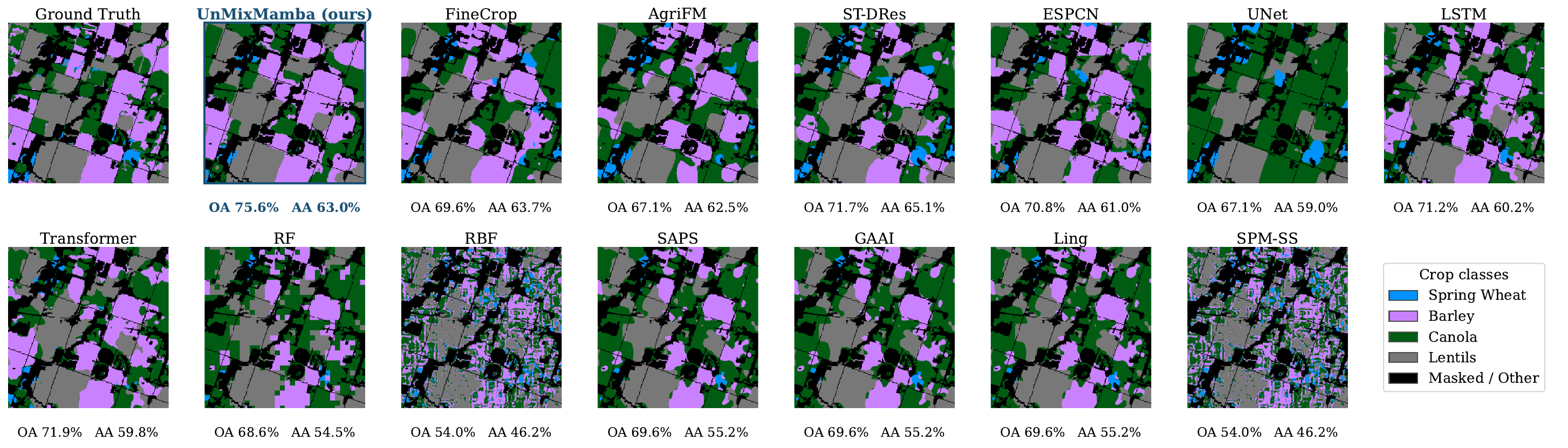}
\vspace{-4pt}
\caption{Barley-focused comparison on an Alberta patch. A $256\times256$ HR window is centered on Barley pixels. Subtitles report patch-level OA and AA; UnMixMamba is highlighted in blue.}
\label{fig:qual_focus_barley}
\end{figure*}

\begin{figure*}[!tp]
\centering
\vspace{-6pt}
\includegraphics[width=\textwidth]{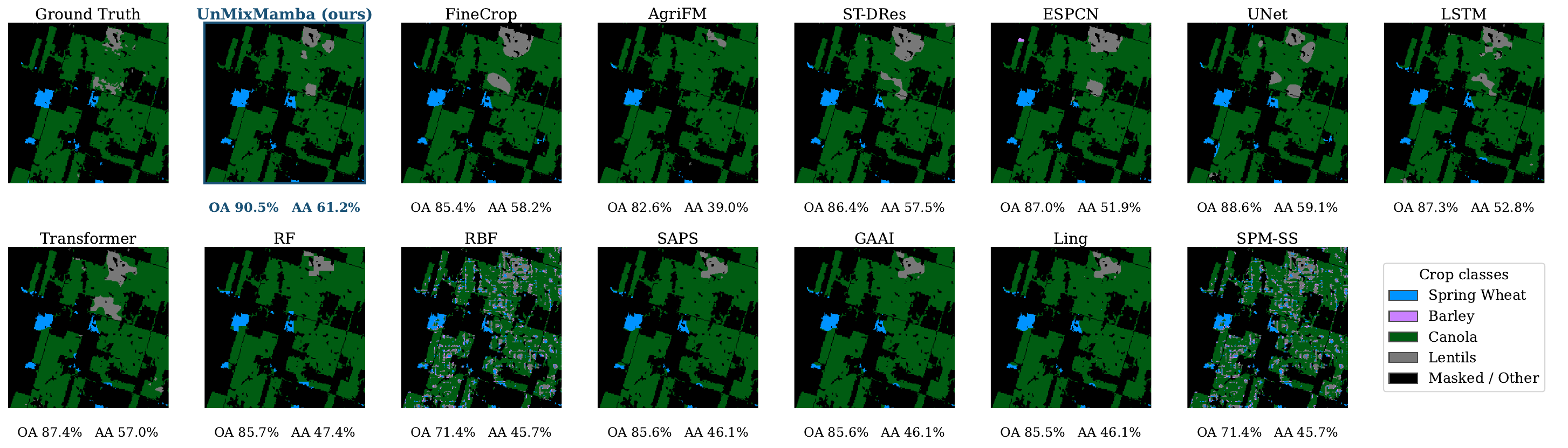}
\vspace{-4pt}
\caption{Canola-focused comparison on an Alberta patch. A $256\times256$ HR window is centered on Canola pixels. Subtitles report patch-level OA and AA; UnMixMamba is highlighted in blue.}
\label{fig:qual_focus_canola}
\end{figure*}

\begin{figure*}[!tp]
\centering
\vspace{-6pt}
\includegraphics[width=\textwidth]{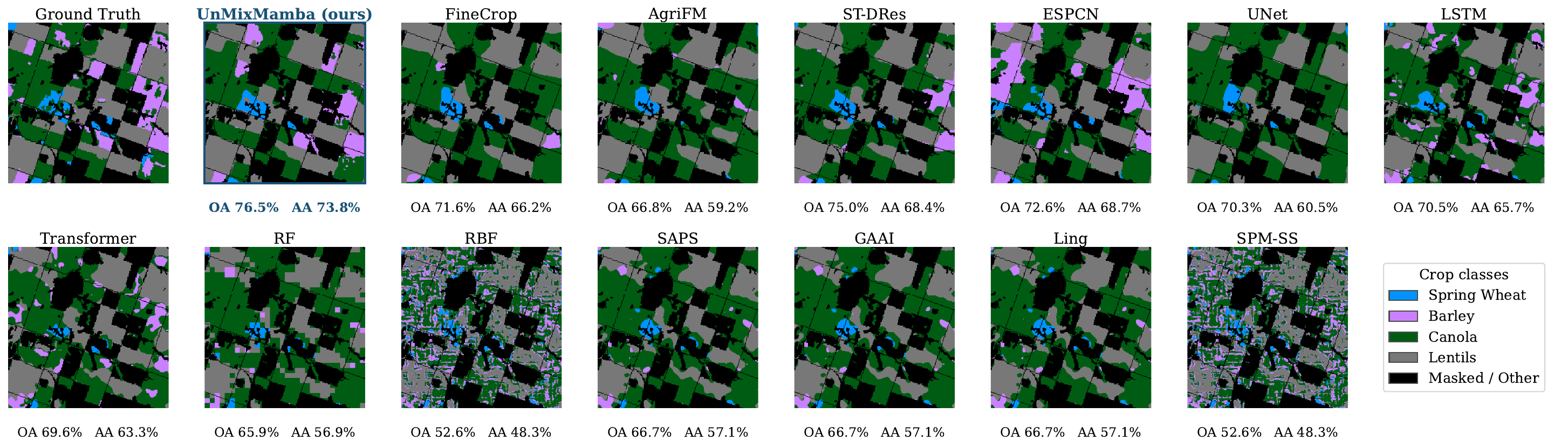}
\vspace{-4pt}
\caption{Lentils-focused comparison on an Alberta patch. A $256\times256$ HR window is centered on Lentils pixels. Subtitles report patch-level OA and AA; UnMixMamba is highlighted in blue.}
\label{fig:qual_focus_lentils}
\end{figure*}

Classical SPM fragments Barley and Lentils into speckle, whereas the direct mapping deep baselines blur cereal boundaries. Fig.~\ref{fig:qualitative_random10} gives a broad visual comparison on ten randomly selected Alberta patches.
Figs.~\ref{fig:qual_focus_sw}--\ref{fig:qual_focus_lentils} then show four class focused Alberta patches. For each crop we center a $256\times256$ HR window on that class and plot AAFC ground truth with the baseline maps in a common layout, marking UnMixMamba with a blue border. Subtitles give patch OA and AA (\%) for every method; the text quotes the strongest baseline (ST-DRes) and the two crop mapping architectures (FineCrop and AgriFM).

\paragraph{Spring Wheat (Fig.~\ref{fig:qual_focus_sw}).}
Deep super resolution baselines oversmooth the fragmented cereal boundaries, while classical SPM methods produce speckled allocations. UnMixMamba preserves continuous field geometry on this Alberta patch (OA 90.7\%), ahead of ST-DRes (88.5\%), FineCrop (88.1\%), and AgriFM (87.1\%).

\paragraph{Barley (Fig.~\ref{fig:qual_focus_barley}).}
Barley is the hardest class in Table~\ref{tab:main_results}. On this Alberta patch UnMixMamba leads with OA 75.6\%, ahead of ST-DRes (71.7\%), FineCrop (69.6\%), and AgriFM (67.1\%), and recovers more coherent elongated Barley parcels than several baselines.

\paragraph{Canola (Fig.~\ref{fig:qual_focus_canola}).}
UnMixMamba maps the Canola-dominated region as contiguous oilseed parcels with stable edges at mixed boundaries (patch OA 90.5\%), ahead of ST-DRes (86.4\%), FineCrop (85.4\%), and AgriFM (82.6\%). Several baselines bleed Canola into adjacent cereal rows.

\paragraph{Lentils (Fig.~\ref{fig:qual_focus_lentils}).}
Lentil fields are often smaller and spectrally mixed with cereals at \SI{250}{m} resolution. On this Alberta patch UnMixMamba leads with OA 76.5\%, ahead of ST-DRes (75.0\%), FineCrop (71.6\%), and AgriFM (66.8\%); classical SPM maps are visibly more fragmented.

\subsection{Ablation and Component Analysis}
\label{sec:ablation}
Table~\ref{tab:all_unmix_experiments} (rows A1--A6) reports a matched 80 epoch leave-one-out study on the Alberta 2015 spatial block test split (183 patches). These rows isolate Contributions~1--3 (blocks C1--C3). Every ``w/o'' row disables or replaces exactly one component under the same training setup; reported drops are relative to UnMixMamba (row P0).

Relative to UnMixMamba, C1 and C3 show large drops under the matched leave-one-out protocol. Replacing the TSSD decoupler with a shallow LR unmixing encoder (w/o C1; a \(1{\times}1\) convolution stack \(D\rightarrow D/2\rightarrow D/4\rightarrow 4\) with batch normalization and ReLU, keeping the stem and channel attention) drops OA by 22.2~pp to 0.6181 and drives Barley recall to 0.0000 while Canola is over predicted (0.9376); the network abandons the hard cereal for the most prevalent class. Removing abundance-guided token selection (w/o C3) shows the opposite pattern: Spring Wheat recall inflates to 0.9464, Barley again falls to 0.0000, and Lentils fall to 0.3491, because uniform allocation spends refinement without abundance budgeting. Keeping only the Barley/Canola token floor without abundance weighting (w/o C3, uniform + floor) still leaves Barley at 0.0239, so the minority floor does not replace mixture-guided budgeting. These collapses indicate that the full pipeline depends on each of C1--C3 rather than on a redundant add-on: without C1 or C3 the remaining modules cannot recover four-crop SPM under this protocol.

Contribution~2 decomposes into complementary parts. Collapsing to one spectrum per crop (w/o C2, single query) costs 6.2~pp OA; Canola stays high (0.918) while Barley falls to 0.152. Bilinear abundance upsampling (w/o C2, bilinear SR) costs 6.6~pp OA and drives Barley to 0.198, isolating ESRT from the variant bank. Zeroing reconstruction losses (w/o C2, no recon) costs 5.8~pp OA with Barley at 0.423, showing that cross-scale spectral consistency regularizes the mixture even when ESRT and the full variant bank remain. The ablation rows are single runs; their OA drops (5.8--22.2~pp) are well above the seed standard deviation of OA (Section~\ref{sec:robustness}).

\subsection{Mixture and Endmember Analysis}
\label{sec:mixture_figs}
The learned endmember bundles (Fig.~\ref{fig:endmembers}) show $R{=}10$ spectral temporal variants per class; Spring Wheat and Barley variants overlap phenologically, consistent with the cereal confusion above, whereas Canola and Lentils retain distinct envelopes. Fig.~\ref{fig:abundance} contrasts the final hard HR map with soft abundance fields on a mixed test patch: fractional Spring Wheat, Barley, Canola, and Lentils responses at field edges drive both cross-scale reconstruction (Eq.~\eqref{eq:recon}) and abundance-guided token budgeting (Eq.~\eqref{eq:token}). Together with the C2 ablations above, this indicates that the mixture-constrained representation is necessary for stable four-crop mapping.

\subsection{Robustness and Sensitivity Analysis}
\label{sec:robustness}
Across three independent 80 epoch retrainings (mean \(\pm\) standard deviation), UnMixMamba averages OA $0.836 \pm 0.004$, AA $0.777 \pm 0.007$, and Kappa $0.767 \pm 0.004$, while ST-DRes under the identical protocol averages OA $0.797 \pm 0.001$, AA $0.706 \pm 0.004$, and Kappa $0.710 \pm 0.001$ (Table~\ref{tab:multiseed_summary}).
UnMixMamba's seed mean is higher on all three metrics, and its small \emph{aggregate} standard deviation ($\pm 0.004$ OA) indicates that overall accuracy is stable across initializations rather than a single favorable seed. Aggregate stability does not imply per-class stability: residual variance is concentrated in the three-way Barley hub. Barley recall ranges from 0.493 to 0.606 across seeds (per-class seed ranges are reported here rather than in Table~\ref{tab:multiseed_summary}) and trades off mainly against Canola (0.845 down to 0.803), with Spring Wheat also involved; i.e., both arms of the Barley confusion oscillate with initialization even though OA barely moves.
This is expected for classes on the decision boundary.

\begin{table}[!t]
\caption{Multi seed summary (80 epoch protocol for UnMixMamba and ST-DRes; three independent seeds per model; mean \(\pm\) standard deviation). Better value in bold.}
\label{tab:multiseed_summary}
\centering
\footnotesize
\setlength{\tabcolsep}{2pt}
\begin{tabular}{lcccc}
\toprule
\textbf{Model} & \textbf{$n$} & \textbf{OA} & \textbf{AA} & \textbf{Kap.} \\
\midrule
UnMixMamba & 3 & \textbf{.836 $\pm$.004} & \textbf{.777 $\pm$.007} & \textbf{.767 $\pm$.004} \\
ST-DRes   & 3 & .797 $\pm$.001 & .706 $\pm$.004 & .710 $\pm$.001 \\
\bottomrule
\end{tabular}
\end{table}

The token budget sweep (rows T1, P0, and T2 of Table~\ref{tab:all_unmix_experiments}), reported as a sensitivity analysis around the default, shows a clear operating regime among sparse settings: the default $\lambda{=}0.10$ gives the best result, raising the budget fivefold to $\lambda{=}0.50$ (still sparse, not dense) lowers OA to 0.8181 and Barley recall to 0.5055 while adding compute, and starving the budget to $\lambda{=}0.05$ loses 8.6~pp OA and collapses Barley recall to 0.0000. A dense baseline would use \(\lambda{=}1\) over all \(262{,}144\) tokens. A fixed \(\lambda\) may not transfer to regions with different abundance distributions.

\subsection{Computational Cost}
\label{sec:complexity}
The cost of the pipeline is dominated by operations on the HR grid, so the two design decisions that control complexity are super-resolving in the low-dimensional abundance space (\(M{=}4\) channels) rather than in the \(138\)-channel radiance space, and budgeting the Mamba refinement. At the default budget \(\lambda{=}0.1\), token selection on the 183 test patches yields on average \(3.0{\times}10^{4}\) unique selected locations per patch (\(\approx 11.4\%\) of the HR grid), with essentially no cross-crop duplication of the selected set (unique/nominal \(=1\)). The average exceeds the nominal 10\% because the Barley/Canola floor (Eq.~\eqref{eq:ntoken}) adds tokens whenever a hard class's abundance-derived count falls below 3\% of the grid. Mean token-budget entropy across the four crops is \(1.28\) (max \(\ln 4{=}1.39\)), and about \(5.6\%\) of selected tokens fall on AAFC label boundaries, so the budget is not primarily an edge detector.

Table~\ref{tab:complexity} reports parameters, estimated GFLOPs, wall-clock time, and accuracy for UnMixMamba and the deep SPM and general-purpose baselines. UnMixMamba at \(\lambda{=}0.1\) uses 6.92\,M parameters, about 1096\,GFLOPs, and 103\,ms per patch on an RTX A6000, versus 0.05\,M, 0.4\,GFLOPs, and 1\,ms for ST-DRes. Relative to ST-DRes, UnMixMamba therefore gains +4.3~pp OA, +9.8~pp mIoU, and the best Barley recall in Table~\ref{tab:main_results} at roughly \(100\times\) the wall-clock time and about \(2700\times\) the GFLOPs. We consider this an acceptable trade-off for offline annual AAFC-style inventory production, not a real-time or edge-deployable alternative to ST-DRes.

\subsection{Geographic Transfer (Saskatchewan)}
\label{sec:cross_province}
To assess geographic generalization beyond the Alberta spatial block benchmark, we apply every Alberta trained model from Table~\ref{tab:main_results} to Saskatchewan 2015, which stresses spatial domain shift in crop geometry and spectral mixtures. All models keep their Alberta weights; evaluation uses the provincial 138 band MODIS 2015 stack and AAFC 2015 labels. We evaluate all 1{,}049 Saskatchewan patches with at least 5\% four-crop coverage; this patch-selection threshold replaces the 10\% load-time filter used for Alberta, and the inference settings are otherwise those of the fair protocol.
The classical SPM allocators use abundance estimates from the Alberta trained RF estimator, as on the Alberta benchmark.

All methods degrade relative to Alberta (Table~\ref{tab:main_results}), reflecting prairie wide domain shift in crop geometry and spectral mixtures. UnMixMamba leads on OA (0.7348), AA (0.6841), Kappa (0.6118), and Barley recall (0.4594). ST-DRes remains close on overall accuracy (0.7253) and shares the best Lentils recall with FineCrop (0.7580). FineCrop and AgriFM reach OA 0.6730 and 0.6732; FineCrop retains more Barley (0.3324), whereas AgriFM attains the best Spring Wheat recall (0.8685). GAAI, SAPS, and Ling reach OA \(\approx 0.715\), above several deep baselines, but only through the highest Canola recall (GAAI 0.8260, column best): like RF, they collapse Barley to $\approx0.08$. RBF and SPM-SS fall furthest (OA 0.5175), and UNet again collapses Barley (0.0000). The Alberta$\rightarrow$Saskatchewan gap confirms that geographic transfer remains challenging.

\subsection{Limitations}
\label{sec:limitations}
UnMixMamba has two main limitations.
\emph{(i)~Barley remains difficult.} Even with UnMixMamba, Barley recall is only 0.55, so almost half of the Barley pixels are still misclassified, mostly as Spring Wheat or Canola. At \SI{250}{m}, Barley looks very similar to these crops over the whole season (Fig.~\ref{fig:modis_signatures}), so separating it more reliably may need additional information, such as finer-resolution imagery.
\emph{(ii)~One year and two provinces.} All experiments use crop year 2015: the model is trained in Alberta and tested in Alberta and Saskatchewan. Its performance in other years has not been tested yet; cross-year evaluation with transfer learning is our next step.

\section{Conclusion}
\label{sec:conclusion}
UnMixMamba targets MODIS-to-AAFC crop subpixel mapping inside a known agricultural mask. The three contributions are (1)~TSSD abundance SPM (C1), (2)~variant-endmember LR/HR reconstruction (C2), and (3)~abundance-guided sparse Mamba refinement (C3). On Alberta 2015, UnMixMamba reaches OA 0.8397, mIoU 0.6640, AA 0.7837, and Kappa 0.7715 on the 183-patch spatial block test split, ahead of ST-DRes and the remaining baselines, with the best Barley recall in the benchmark. Multi-seed retraining of the full system is stable in aggregate OA, while leave-one-out ablations show that C1--C3 are interdependent rather than redundant. Under geographic transfer to Saskatchewan 2015, UnMixMamba keeps the lead in OA, AA, Kappa, and Barley recall. Relative to ST-DRes, the +4.3~pp OA / +9.8~pp mIoU gain costs roughly \(100\times\) wall-clock time (103\,ms vs.\ 1\,ms; 1096\,GFLOPs at \(\lambda{=}0.1\)), which we consider acceptable for offline annual mapping. Our next step is cross-year evaluation through transfer learning; further work includes independent crop/non-crop masks or an explicit background class, shift and PSF sensitivity analysis, mixed-pixel-aware and adaptive token budgets, attribution of endmember variants to ancillary drivers, and backbone compression.
\FloatBarrier
\section*{Acknowledgment}
The authors thank Yuxian Wang and colleagues for providing the ST-DRes implementation used in this study \cite{stdres2022}.

\bibliographystyle{IEEEtran}
\bibliography{TGRS_references}

\end{document}